\documentclass{article}
\usepackage{ijcai26}

\usepackage{times}
\usepackage{soul}
\usepackage{url}
\usepackage[hidelinks]{hyperref}
\usepackage[utf8]{inputenc}
\usepackage[small]{caption}
\usepackage{graphicx}
\usepackage{amsmath}
\usepackage{amsthm}
\usepackage{booktabs}
\usepackage{algorithm}
\usepackage{algorithmic}
\usepackage[switch]{lineno}
\usepackage{amsfonts}

\newcommand{\reffig}[1]{Fig.~\ref{#1}}
\newcommand{\reftable}[1]{Table \ref{#1}}
\usepackage{booktabs, multirow, caption}
\usepackage{booktabs}
\usepackage{multirow}
\usepackage{amssymb} 
\usepackage{graphicx}

\newcommand{\INPUT}{\REQUIRE} 

\title{FMVP: Masked Flow Matching for Adversarial Video Purification}

\author{
Duoxun Tang$^1$
\and
Xueyi Zhang$^2$\and 
Chak Hin Wang$^{3}$\and
Xi Xiao $^{1,*}$\\
Dasen Dai $^{4}$\and
Xinhang Jiang$^2$\and
Wentao Shi$^1$\and
Rui Li$^5$\And
Qing Li$^5$
\affiliations
$^1$Tsinghua University, China\\
$^2$The Chinese University of Hong Kong, Shenzhen, China\\
$^3$University of New South Wales, Australia,
$^4$The Chinese University of Hong Kong, China\\
$^5$Peng Cheng Laboratory, Shenzhen, China\\
\emails
\{tdx25, shiwt25\}@mails.tsinghua.edu.cn,
1155211130@link.cuhk.edu.hk, xinhangjiang@link.cuhk.edu.cn,
zhangxy1998@163.com, chak\_hin.wang@student.unsw.edu.au\\ 
$^*$ Corresponding author: \texttt{xiaox@sz.tsinghua.edu.cn}
}

\begin{document}

\maketitle

\begin{abstract}
Video recognition models remain vulnerable to adversarial attacks, while existing diffusion-based purification methods suffer from inefficient sampling and curved trajectories. Directly regressing clean videos from adversarial inputs often fails to recover faithful content due to the subtle nature of perturbations; this necessitates physically shattering the adversarial structure. Therefore, we propose \underline{F}low \underline{M}atching for Adversarial \underline{V}ideo \underline{P}urification ($\mathbf{FMVP}$). FMVP physically shatters global adversarial structures via a masking strategy and reconstructs clean video dynamics using Conditional Flow Matching (CFM) with an inpainting objective. To further decouple semantic content from adversarial noise, we design a Frequency-Gated Loss (FGL) that explicitly suppresses high-frequency adversarial residuals while preserving low-frequency fidelity. We design Attack-Aware and Generalist training paradigms to handle known and unknown threats, respectively. Extensive experiments on UCF-101 and HMDB-51 demonstrate that FMVP outperforms state-of-the-art methods (DiffPure, Defense Patterns (DP), Temporal Shuffling (TS) and FlowPure), achieving robust accuracy exceeding 87\% against PGD and 89\% against CW attacks. Furthermore, FMVP demonstrates superior robustness against adaptive attacks (DiffHammer) and functions as a zero-shot adversarial detector, attaining AUC-ROC scores of 0.98 for PGD and 0.79 for highly imperceptible CW attacks.
\end{abstract}

\section{Introduction}

Adversarial attacks~\cite{su2019one,zhang2025adversarial} pose a critical threat to deep neural networks (DNNs) in video recognition tasks ~\cite{ji20123d,wang2024multimodal,ijcai2025p276,tang2025video}, despite their remarkable success. These attacks involve inputs modified by imperceptible perturbations~\cite{madry2017towards,carlini2017towards} to induce misclassification. This vulnerability poses severe security risks in safety-critical applications such as autonomous driving \cite{xu2021action}, medical imaging diagnosis~\cite{abdou2022literature,ijcai2025p60} and facial recognition systems \cite{wang2021deep,li2025towards}. In the video domain, high dimensionality and temporal redundancy create a vast attack surface, allowing adversaries to craft potent attacks using gradient-based methods like Projected Gradient Descent (PGD)~\cite{madry2017towards} and optimization-based methods such as Carlini \& Wagner (CW) ~\cite{carlini2017towards}, and powerful adaptive attack methods such as DiffHammer (DH)~\cite{wang2024diffhammer}. Traditional defenses such as Adversarial Training ~\cite{madry2017towards,gowal2020uncovering,wang2023better,singh2023revisiting}, incur prohibitive computational costs and often degrade performance on clean data, making them impractical for large-scale video models~\cite{zhang2023video,lin2024video}.

\begin{figure}[t]
  \includegraphics[width=1\linewidth]{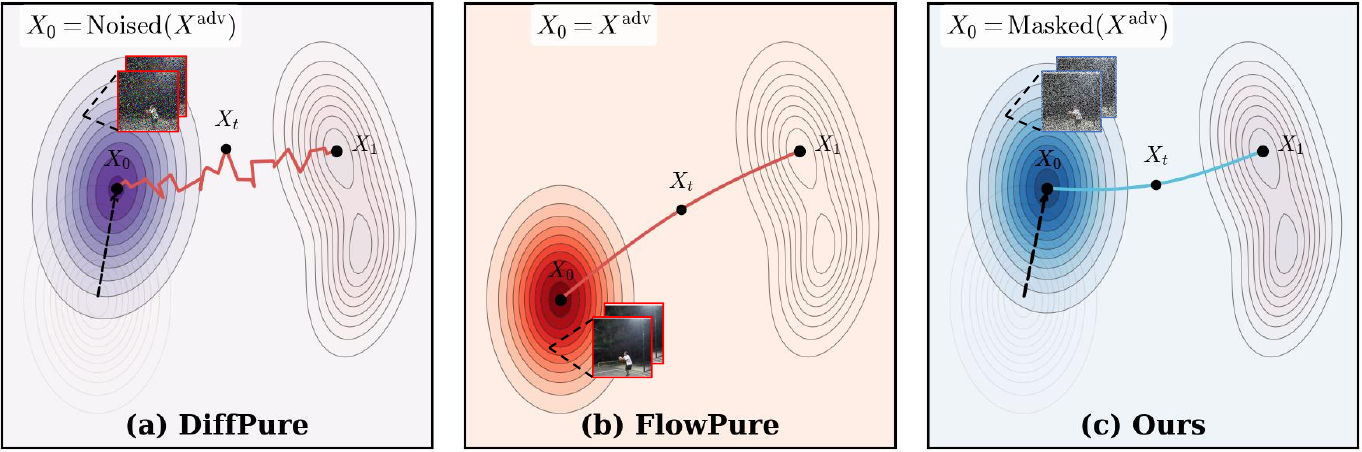}
\vspace{-3mm}
  \caption{During inference, unlike DiffPure (a) which shifts the distribution via noise injection (purple)  and FlowPure (b) which initiates directly from the original adversarial distribution (red), ours (c) employs masking to physically disrupt adversarial patterns. This shifts the input to a specific masked adversarial distribution (blue), effectively shattering the attack structure while preserving the original semantics via inpainting-based reconstruction.}
\label{figs:Abstract}
\end{figure}

To address these limitations, \textit{adversarial purification}~\cite{pouya2018defense,samangouei2018defense,yoon2021adversarial,nie2022diffusion,lee2023defending,hwang2024temporal,collaert2025flowpure} has emerged as a mainstream defense strategy, aiming to remove perturbations from inputs prior to inference without modifying the classifier. Video-specific heuristics, such as Defense Patterns (DP) \cite{lee2023defending} and Temporal Shuffling (TS) \cite{hwang2024temporal}, rely on input transformations to obfuscate adversarial gradients. However, these methods essentially rely on gradient masking without projecting data back to the clean manifold, thereby failing to recover semantically valid inputs. Recent state-of-the-art purification approaches are largely dominated by generative models, specifically Diffusion-based methods such as DiffPure~\cite{nie2022diffusion}, which purify inputs via a stochastic forward-reverse diffusion process. Alternatively, FlowPure~\cite{collaert2025flowpure} is a method based on Continuous Normalizing Flows (CNFs)~\cite{chen2018neural} trained with Conditional Flow Matching (CFM)~\cite{lipman2022flow,liu2022flow}  to map adversarial examples to clean counterparts. DiffPure's reverse diffusion process during inference aims to disrupt adversarial patterns by first diffusing the input to a noisy state and then denoising it back, but it suffers from inefficient sampling and stochastic processes with random, curved trajectories (\reffig{figs:Abstract} (a)), while FlowPure addresses these through deterministic Ordinary Differential Equations (ODE) integration along straighter paths (\reffig{figs:Abstract} (b)). However, directly modeling the transition from adversarial video to clean video may lead to lazy learning, as the subtle differences between closely located samples make it difficult for the model to push adversarial inputs away from the nearby adversarial manifold, resulting in suboptimal purification performance. Therefore, additional mechanisms are needed to actively disrupt adversarial structures and guide the model toward clean reconstruction. Moreover, adversarial perturbations typically manifest as high-frequency anomalies in the spectral domain, distinct from the low-frequency dominance of semantic video content. In contrast, existing purification methods largely overlook this spectral energy distribution, focusing solely on spatial reconstruction constraints.

To fill the gaps mentioned above, this paper proposes a novel purification method named FMVP (Flow Matching for Adversarial Video Purification) that first disrupts the adversarial patterns in adversarial videos through masking (\reffig{figs:Abstract} (c)). The mask at an appropriate ratio can not only destroy adversarial patterns but also preserve adjacent pixels, enhancing the model's ability for semantic reconstruction. Moreover, in velocity field prediction, an additional Fast Fourier Transform (FFT) is applied to construct a Frequency-Gated Loss (FGL) that explicitly suppresses high-frequency adversarial noise while preserving the low-frequency semantic fidelity of the video. We explore two training paradigms: an Attack-Aware version tailored to rectify specific perturbations (e.g., from PGD or CW), and a generalist version trained with gaussian noise to handle unknown threats. Extensive experiments on UCF-101 \cite{soomro2012ucf101} and HMDB-51 ~\cite{kuehne2011hmdb} demonstrate that FMVP achieves robust  accuracy exceeding 87\% against PGD and over 89\% against CW attacks, and outperforms SOTA methods such as DiffPure-DDPM, DiffPure-DDIM, DP, TS, and FlowPure. FMVP achieves better robustness against adaptive attacks (DH) while functioning as a effective adversarial detector, achieving AUC-ROC scores of 0.98 for PGD and 0.79 for highly imperceptible CW attacks. 

Our key contributions include:
\begin{enumerate}
    \item A novel framework Flow Matching for Adversarial Video Purification (FMVP) is proposed to disrupt adversarial patterns and purify videos by integrating Conditional Flow Matching with masking.
    
    \item A Frequency-Gated Loss (FGL) is designed based on spectral analysis, acting as a soft gate in velocity field prediction that preserves low-frequency semantics while suppressing high-frequency adversarial noises.
    
    \item Extensive experiments are conducted on UCF-101 and HMDB-51, showing that FMVP outperforms state-of-the-art methods under standard (PGD and CW) and adaptive (DH) attacks and functions effectively as an adversarial detector.
    
\end{enumerate}

\begin{figure*}[t]
\begin{center}
  \includegraphics[width=0.95\linewidth]{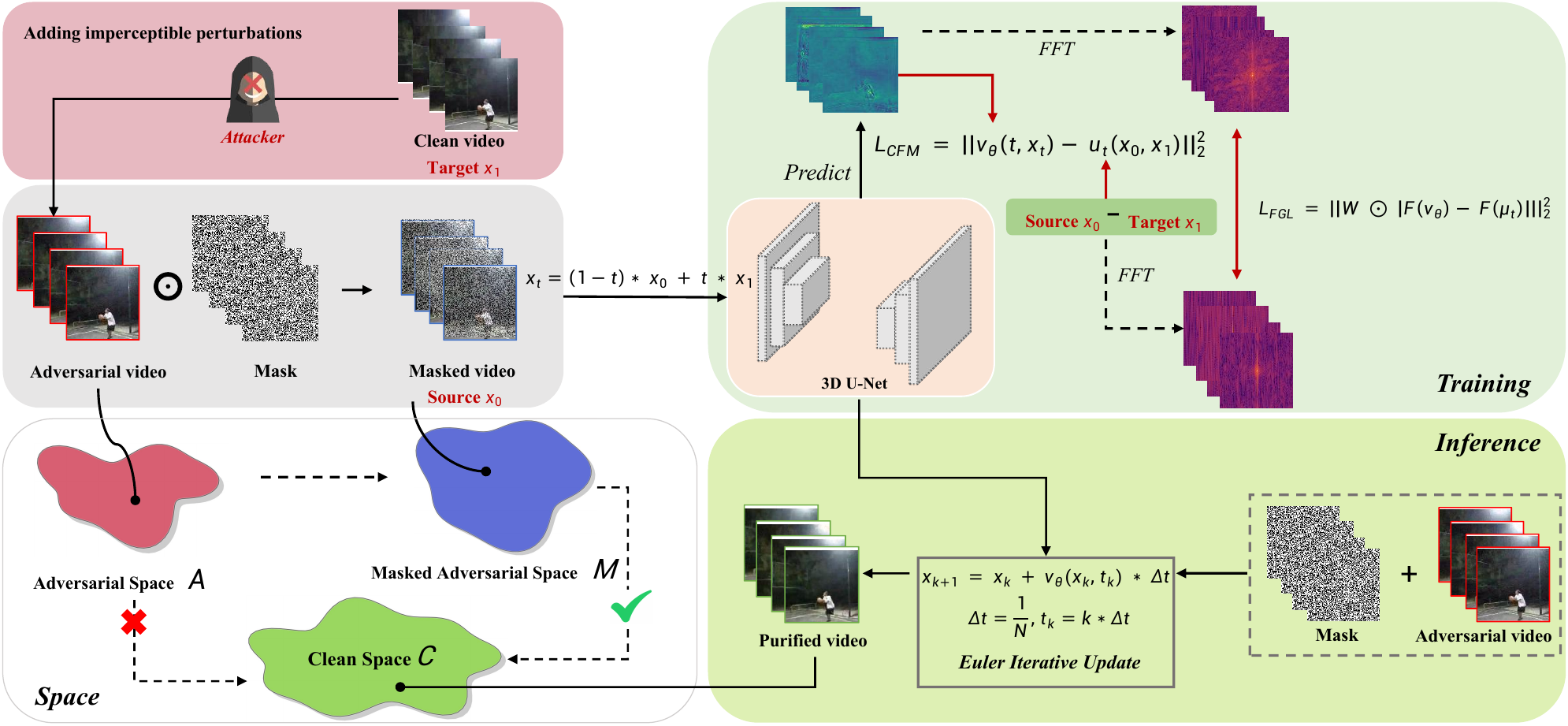}
\end{center}
\vspace{-3mm}
  \caption{Overview of FMVP. Adversarial videos are transformed from the Adversarial Space $\mathcal{A}$ to the Masked Adversarial Space $\mathcal{M}$. This process disrupts adversarial patterns while preserving the original semantics of adjacent pixels. The training phase is conducted under the constraints of the Mean Squared Error (MSE) loss and the Frequency-Gated Loss within the Conditional Flow Matching (CFM) framework, while inference employs the Euler iterative update.}
\label{figs:FMVP_pipline}
\end{figure*}

\section{Related Work}
\subsection{Diffusion Models and Flow Matching}

Denoising Diffusion Probabilistic Models (DDPMs)~\cite{ho2020denoising} function by reversing a forward diffusion process that progressively corrupts data $\mathbf{x}_0 \sim q(\mathbf{x})$ into Gaussian noise. The forward transition kernel $q(\mathbf{x}_t|\mathbf{x}_0)$ allows for the direct sampling of latent variable $\mathbf{x}_t$ at arbitrary timestep $t \in [0, T]$:
\begin{equation}
    q(\mathbf{x}_t|\mathbf{x}_0) = \mathcal{N}(\mathbf{x}_t; \sqrt{\bar{\alpha}_t}\mathbf{x}_0, (1-\bar{\alpha}_t)\mathbf{I}),
\end{equation}
where $\bar{\alpha}_t$ is the noise schedule. The generative reverse process $p_\theta(\mathbf{x}_{t-1}|\mathbf{x}_t)$ is parameterized by a neural network $\boldsymbol{\epsilon}_\theta(\mathbf{x}_t, t)$ trained via the simplified variational lower bound objective:
\begin{equation}
    \mathcal{L}_{\text{DDPM}} = \mathbb{E}_{t, \mathbf{x}_0, \boldsymbol{\epsilon}}\left[ \| \boldsymbol{\epsilon} - \boldsymbol{\epsilon}_\theta(\sqrt{\bar{\alpha}_t}\mathbf{x}_0 + \sqrt{1-\bar{\alpha}_t}\boldsymbol{\epsilon}, t) \|^2 \right].
\end{equation}
To accelerate sampling, Denoising Diffusion Implicit Models (DDIMs)~\cite{song2020denoising} generalize the Markovian process to a non-Markovian deterministic mapping. While diffusion models rely on stochastic chains or SDEs, Flow Matching (FM) trains Continuous Normalizing Flows (CNFs) by regressing a time-dependent vector field $v_t$ that generates a probability path $p_t$ satisfying the continuity equation $\frac{\partial p_t}{\partial t} + \nabla \cdot (p_t v_t) = 0$. The flow is defined by the ODE:
\begin{equation}
    \frac{d\mathbf{x}}{dt} = v_t(\mathbf{x}), \quad \text{s.t.} \quad \mathbf{x}_0 \sim p_0, \mathbf{x}_1 \sim p_{data}.
\end{equation}
To circumvent the intractability of the marginal vector field, Conditional Flow Matching (CFM) minimizes the regression loss over conditional flows $u_t(\mathbf{x}|\mathbf{z})$:
\begin{equation}
    \mathcal{L}_{\text{CFM}}(\theta) = \mathbb{E}_{t, q(\mathbf{z}), p_t(\mathbf{x}|\mathbf{z})} \left[ \| v_\theta(\mathbf{x}, t) - u_t(\mathbf{x}|\mathbf{z}) \|^2 \right].
\end{equation}
The most efficient variant utilizes Optimal Transport (OT)~\cite{liu2022flow} displacement paths. Given a source sample $\mathbf{x}_0 \sim \mathcal{N}(\mathbf{0}, \mathbf{I})$ and target $\mathbf{x}_1 \sim p_{data}$, the conditional probability path $\mu_t$ and the target conditional vector field $u_t$ are rigorously defined as linear interpolations:
\begin{equation}
    \mu_t(\mathbf{x}_0, \mathbf{x}_1) = (1 - t)\mathbf{x}_0 + t\mathbf{x}_1, \quad u_t(\mathbf{x}|\mathbf{x}_0, \mathbf{x}_1) = \mathbf{x}_1 - \mathbf{x}_0.
\end{equation}

\subsection{Adversarial Purification}

Current leading purification strategies are primarily built upon generative modeling, with diffusion-based approaches setting the standard. DiffPure~\cite{nie2022diffusion}, for example, purifies adversarial inputs by first perturbing them slightly through the forward diffusion process and then reconstructing clean samples via reverse-time stochastic dynamics. As a deterministic counterpart, FlowPure~\cite{collaert2025flowpure} replaces stochastic sampling with continuous normalizing flows trained under CFM, enabling direct and efficient mapping of corrupted inputs back to the clean data manifold. Beyond image-level defenses, video-specific methods attempt to exploit temporal structure for robustness. Temporal Shuffling (TS) ~\cite{hwang2024temporal} disrupts adversarial optimization by randomly reordering frames, thereby breaking gradient coherence across time. Defense Patterns (DP) ~\cite{lee2023defending}, on the other hand, overlays fixed spatial masks onto input sequences to obscure perturbations. Despite their use of domain-specific priors, these video defenses often rely on ad hoc transformations or heavy randomness, limiting their generalization and purification fidelity.

\subsection{Research Gap} 
Existing purification methods face several unresolved challenges. (1): Diffusion-based approaches often exhibit instability due to the inherent stochasticity of the generative process. (2): While FlowPure introduced CNFs for image purification, it relies on a direct mapping without structural disruption mechanisms. In the video domain where adversarial patterns are temporally coherent across frames, direct mapping typically fails to thoroughly eliminate perturbations because the model might lazily preserve the adversarial structure to minimize reconstruction loss. (3): Prior works largely overlook frequency domain constraints. They neglect to regularize learning toward perceptually meaningful reconstructions and fail to suppress high-frequency components that often carry or amplify adversarial patterns.

\section{Methodology}
\subsection{Preliminary}

Let $f_\phi$ denote a pre-trained video classifier parameterized by $\phi$, and $(\mathbf{x}^{\mathrm{clean}}, y)$ represent a clean video sample and its corresponding ground-truth label. An adversarial attack algorithm $\mathcal{A}$ generates an adversarial video $\mathbf{x^{adv}} = \mathcal{A}(\mathbf{x}^{\mathrm{clean}}, y, f_\phi)$ by adding an imperceptible perturbation, such that the classifier is misled into making an incorrect prediction, \textit{i.e.}, $f_\phi(\mathbf{x^{adv}}) \neq y$. Generation-based adversarial purification aims to learn a generator $G_\theta$ that maps the adversarial video $\mathbf{x^{adv}}$ back to the clean data manifold. The objective is to remove the adversarial perturbations while preserving semantic content, ensuring that the purified video is correctly classified by the target model: $f_\phi(G_\theta(\mathbf{x^{adv}})) = y$.

\begin{figure}[t]
  \includegraphics[width=1\linewidth]{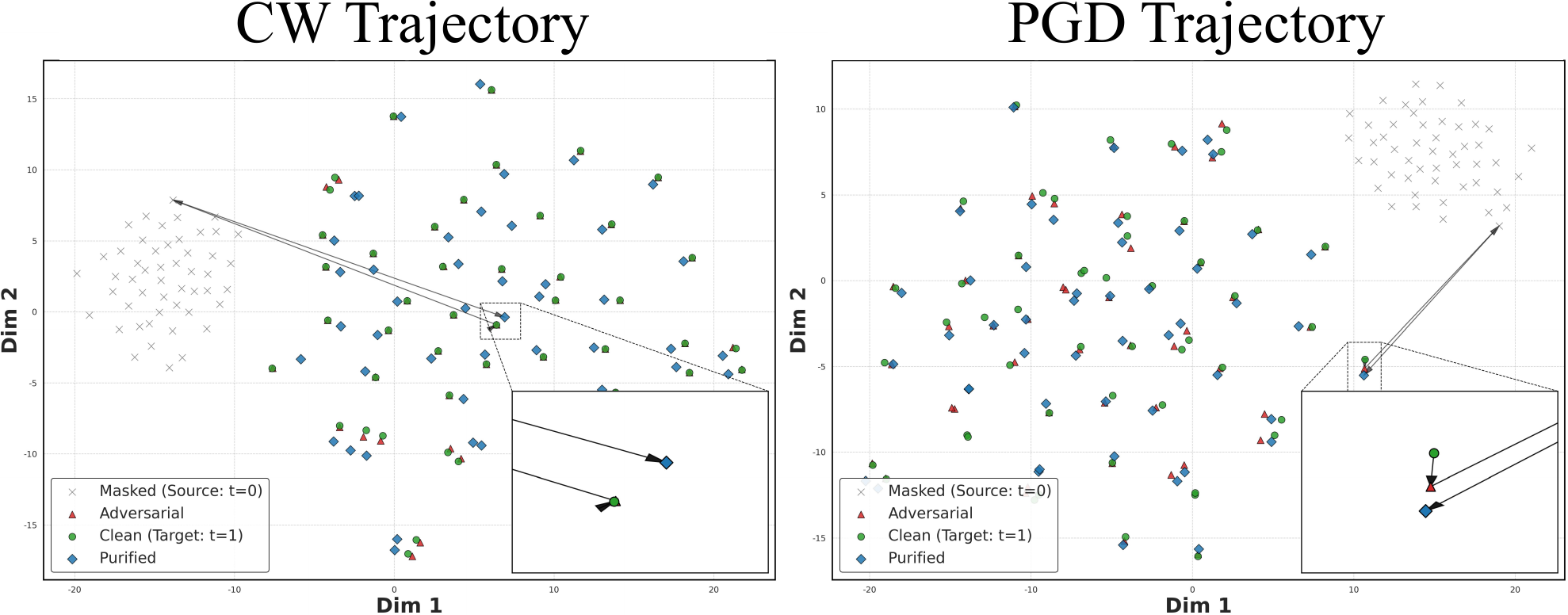}
\vspace{-3mm}
  \caption{The trajectories of 50 real samples in the space from clean $\rightarrow$ adversarial $\rightarrow$ masked $\rightarrow$ purified. Left: CW attack; right: PGD attack.}
\label{figs:State_visual}
\end{figure}

\subsection{Masked Conditional Flow Matching}
\reffig{figs:FMVP_pipline} illustrates the overview of FMVP. Let $\mathbf{x}^{\mathrm{adv}} \in \mathbb{R}^{B \times C \times T \times H \times W}$ denote an adversarially perturbed video input. Adversarial perturbations, such as those generated by PGD or CW attacks, are often imperceptibly small in pixel space. 
As shown in \reffig{figs:State_visual}, both attacks are highly imperceptible. And CW is especially close to the original clean video. This highlights the necessity of disrupting such adversarial patterns so that the purified video can be restored to a neighborhood of the clean data manifold. Direct purification from $\mathbf{x}^{\mathrm{adv}}$ to the clean target $\mathbf{x}^{\mathrm{clean}}$ may lead to lazy learning, wherein the model preserves residual adversarial patterns while minimizing superficial pixel-wise error, thereby hindering its ability to predict a velocity field capable of effectively removing the perturbations. To mitigate this, we introduce a stochastic masking mechanism that disrupts coherent adversarial structures before applying CFM.

Unlike traditional zero-filling, which introduces artificial discontinuities and distribution shifts, filling masked regions with Gaussian noise aligns the input with the standard source distribution inherent to the Flow Matching paradigm~\cite{lipman2022flow}. This strategy effectively unifies spatial inpainting with generative denoising, allowing the model to leverage its learned priors to reconstruct semantically consistent content from the noise~\cite{lugmayr2022repaint}. Specifically, we construct a source sample $\mathbf{x}_0$ by blending $\mathbf{x}^{\mathrm{adv}}$ with standard Gaussian noise under a random binary mask $\mathbf{m} \in \{0,1\}^{B \times C \times T \times H \times W}$:
\begin{equation}
    \mathbf{x}_0 = \mathbf{m} \odot \mathbf{x}^{\mathrm{adv}} + (1 - \mathbf{m}) \odot \boldsymbol{\epsilon}, \quad \boldsymbol{\epsilon} \sim \mathcal{N}(0, \mathbf{I}),
\end{equation}
where $\odot$ denotes element-wise multiplication. The mask $\mathbf{m}$ is generated per sample by first sampling a keep ratio $\rho \sim \mathcal{U}(0.2, 0.6)$, then setting $\mathbf{m}_{b,c,t,h,w} = 1$, if a uniform random value at that location is less than $\rho$, and $0$ otherwise. This strategy partially preserves clean content while injecting unstructured noise into the remainder, thereby breaking adversarial pattern. 

Given $\mathbf{x}_0$ and the target $\mathbf{x}_1 := \mathbf{x}^{\mathrm{clean}}$, We follow the design principles of Rectified Flows~\cite{liu2022flow}:
\begin{equation}
    \mathbf{x}_t = (1 - t) \mathbf{x}_0 + t \mathbf{x}_1, \quad t \sim \mathcal{U}(0, 1).
\end{equation}
The ground-truth velocity field along this path is constant:
\begin{equation}
    \mathbf{u}^\star = \frac{d\mathbf{x}_t}{dt} = \mathbf{x}_1 - \mathbf{x}_0.
\end{equation}
Our network $v_\theta(\cdot, t)$ learns to predict this velocity conditioned on time $t$ and the current state $\mathbf{x}_t$. The core CFM objective minimizes the discrepancy between predicted and true velocities:
\begin{equation}
    \mathcal{L}_{\mathrm{CFM}}(\theta) = \mathbb{E}_{t, \mathbf{x}_0, \mathbf{x}_1} \left[ \left\| v_\theta(\mathbf{x}_t, t) - (\mathbf{x}_1 - \mathbf{x}_0) \right\|_2^2 \right].
\end{equation}
Noting that our framework trains on three variants: (i) PGD-based ($FMVP^{PGD}$), (ii) CW-based ($FMVP^{CW}$), and (iii) attack-agnostic Gaussian masking ($FMVP^{Gaussian}$),  
where the first two are trained on $\mathbf{x}^{\mathrm{adv}}$ generated by known attacks (PGD and CW, respectively), while $\mathbf{x}_{\mathrm{0}}$ in the third variant is created by masking $\mathbf{x}^{\mathrm{clean}}$ with Gaussian noise without assuming knowledge of the attack type.

\subsection{Frequency-Gated Reconstruction Constrain}
To further regularize learning toward perceptually meaningful reconstructions and explicitly suppress high-frequency components that often carry or amplify adversarial patterns, we introduce a frequency-domain constrain. Let $\mathcal{F}(\cdot)$ denote the 2D real-valued discrete Fourier transform (RDFT) applied independently to each channel, temporal frame, and batch element. For an input tensor $\mathbf{y} \in \mathbb{R}^{H \times W}$, the RDFT yields a complex-valued spectrum $\widehat{\mathbf{Y}} = \mathcal{F}(\mathbf{y}) \in \mathbb{C}^{H \times W'}$, where $W' = \lfloor W/2 \rfloor + 1$, defined as:
\begin{equation}
    \widehat{Y}_{k,\ell} = \frac{1}{\sqrt{HW}} \sum_{m=0}^{H-1} \sum_{n=0}^{W-1} y_{m,n} \, e^{-2\pi i \left( \frac{km}{H} + \frac{\ell n}{W} \right)},
\end{equation}
for $k = 0, \dots, H-1$ and $\ell = 0, \dots, W'-1$. The orthonormal scaling factor $1/\sqrt{HW}$ ensures energy preservation under the Parseval identity. Due to the conjugate symmetry of the Fourier transform of real-valued signals, only the non-redundant half-spectrum (including the Nyquist frequency when $W$ is even) is retained, which reduces computational overhead while preserving full reconstructability.

\begin{figure}[t]
  \includegraphics[width=0.95\linewidth]{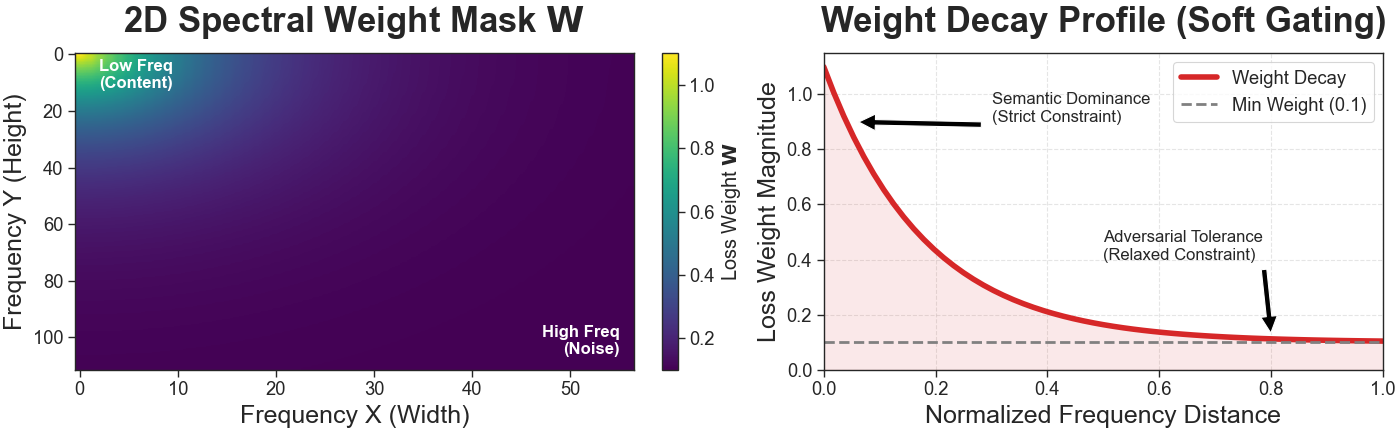}
\vspace{-3mm}
  \caption{Visualization of the Frequency-Gated Loss properties. (Left) The 2D spectral weight mask $\mathbf{W}$ shows that high weights concentrate in the low-frequency center. (Right) The 1D decay profile demonstrates the exponential drop in importance as frequency increases.}
\label{figs:FGLoss}
\end{figure}

The magnitude difference is computed in the frequency domain between the predicted velocity field $v_\theta(\mathbf{x}_t, t)$ and the target $(\mathbf{x}_1 - \mathbf{x}_0)$. To emphasize low-frequency fidelity, where semantic content predominantly resides, we construct a dynamic weight map based on the normalized distance from the DC component (top-left corner of the RDFT output). Let $h = H$ and $w_f = \lfloor W/2 \rfloor + 1$ be the spatial height and frequency-domain width after RDFT. Define normalized coordinate grids:
\begin{equation}
\begin{split}
    y_i &= \frac{i}{h}, \quad i = 0, 1, \dots, h-1, \\
    x_j &= \frac{j}{w_f}, \quad j = 0, 1, \dots, w_f-1.
\end{split}
\end{equation}
The normalized Euclidean distance to the origin is:
\begin{equation}
    d_{ij} = \sqrt{ y_i^2 + x_j^2 }, \quad \forall i \in [0, h), \, j \in [0, w_f).
\end{equation}
We then define a gating function that decays exponentially with distance:
\begin{equation}
    w_{ij} = \exp(-\tau \cdot d_{ij}) + 0.1,
\end{equation}
ensuring low frequencies receive near-unit weight while high frequencies are downweighted but not eliminated, where the exponential decay rate is controlled by $\tau = 5$ (the $+0.1$ floor prevents gradient vanishing). The frequency-gated loss is:
\begin{equation}
    \mathcal{L}_{\mathrm{FGL}}(\theta) = \mathbb{E} \left[ \left\| \mathbf{W} \odot \left| \mathcal{F}(v_\theta(\mathbf{x}_t, t)) - \mathcal{F}(\mu_t(\mathbf{x}_0, \mathbf{x}_1)) \right| \right\|_2^2 \right],
\end{equation}
where $|\cdot|$ denotes complex magnitude, $\mathbf{W}$ is the broadcasted weight tensor with entries $w_{ij}$ replicated across $B$, $C$, and $T$, and all operations are element-wise. \reffig{figs:FGLoss} visualizes the spatial distribution and decay profile of the generated weight mask. It explicitly demonstrates how our Frequency-Gated Loss decouples signal from adversarial noise, acting as a spectral barrier that prevents the model from overfitting to high-frequency adversarial patterns. Overall, the total training loss is given by
\begin{equation}
    \mathcal{L}_{\mathrm{total}}(\theta) = \lambda_{\mathrm{CFM}} \, \mathcal{L}_{\mathrm{CFM}}(\theta) + \lambda_{\mathrm{FGL}} \, \mathcal{L}_{\mathrm{FGL}}(\theta),
\end{equation}
where $\lambda_{\mathrm{CFM}} = 1$ and $\lambda_{\mathrm{FGL}} = 0.2$ balance pixel-space flow alignment and low-frequency structural fidelity to enhance robustness against adversarial perturbations.

\subsection{Inference via Masked Euler Purification}
During inference, we use the same masking strategy that treats purification as an inpainting task. We first construct a hybrid source state $\mathbf{x}_0 = \mathbf{m} \odot (\mathbf{x}_{adv} + \xi \boldsymbol{\epsilon}) + (1 - \mathbf{m}) \odot \boldsymbol{\epsilon}$, where $\xi$ is a negligible noise factor (e.g., $10^{-5}$) introduced to ensure numerical stability by preventing distribution degeneracy in the retained regions, $\mathbf{m}$ is a binary mask sampled with ratio $\gamma$ and $\boldsymbol{\epsilon} \sim \mathcal{N}(\mathbf{0}, \mathbf{I})$. This initialization physically disrupts the global coherence of adversarial perturbations while retaining partial semantic context. Subsequently, we recover the clean video by solving the probability flow ODE via Euler discretization: $\mathbf{x}_{{k+1}} = \mathbf{x}_{{k}} + v_\theta(\mathbf{x}_k, t_k) \cdot \Delta t$, advancing from $t_k=0$ to $1$. The final output ${\mathbf{x^{purified}}}$ is obtained by clamping $\mathbf{x}_1$ to the valid pixel range.

\begin{table*}[t]
\centering

\renewcommand{\arraystretch}{0.2}

\setlength{\tabcolsep}{2pt}

\begin{tabular}{l|cccc|cccc|c}
\toprule
\multirow{2}{*}{\textbf{Method}} 
  & \multicolumn{4}{c|}{\textbf{UCF-101}} 
  & \multicolumn{4}{c|}{\textbf{HMDB-51}} 
  & \multirow{2}{*}{\textbf{Avg. Robust}} \\
\cmidrule(lr){2-5} \cmidrule(lr){6-9}
  & \textbf{Clean} & \textbf{Robust} & \textbf{SSIM} & \textbf{PSNR}
  & \textbf{Clean} & \textbf{Robust} & \textbf{SSIM} & \textbf{PSNR} & \\
\midrule
\multicolumn{10}{l}{\textit{PGD Attack ($\ell_\infty$, $\epsilon=8/255$):}} \\
\hspace{1em}DiffPure-DDPM \hspace{0.2em}\cite{nie2022diffusion}   & 84.0 & 70.0 & 0.8256 & 28.0512 & 89.0 & 75.0 & 0.8364 & 29.0245 & 72.5 \\
\hspace{1em}DiffPure-DDIM \hspace{0.5em}\cite{nie2022diffusion}   & 89.0 & 72.0 & 0.8155 & 28.5201 & 93.0 & 84.0 & 0.8531 & 30.2210 & 78.0 \\
\hspace{1em}DP  \hspace{4.9em}\cite{lee2023defending}   & 94.0 & 56.0 & 0.8758 & 28.8415 & 96.0 & 51.0 & 0.8514 & 29.6124 & 53.5 \\
\hspace{1em}TS  \hspace{4.55em}\cite{hwang2024temporal}   & 92.0 & 74.0 & \underline{0.9437} & 25.6626 & 88.0 & 71.0 & \underline{0.9271} &  24.3389 & 72.5 \\
\hspace{1em}FlowPure \hspace{1.15em}\cite{collaert2025flowpure}     & 94.0 & 77.0 & 0.8815 & 29.0451 & 96.0 & 81.0 & 0.8711 & 29.6333 & 79.0 \\
\hspace{1em}\textbf{FMVP$^{\mathrm{CW}}$} \hspace{2.2em}(Ours)      & 96.0 & 72.0 & 0.8718 & 29.5671 & \underline{97.0} & 85.0 & 0.8779 & 30.1256 & 78.5 \\
\hspace{1em}\textbf{FMVP$^{\mathrm{PGD}}$} \hspace{1.80em}(Ours)      & 95.0 & \underline{84.0} & 0.8896 & \underline{29.6415} & 94.0 & \underline{91.0} & 0.8812 & 29.5462 & \underline{87.5} \\
\hspace{1em}\textbf{FMVP$^{\mathrm{Gaussian}}$}  \hspace{0.4em}(Ours)  & \underline{96.0} & 78.0 & 0.8942 & 28.1649 & 94.0 & 89.0 & 0.8881 & \underline{30.2247} & 83.5 \\

\midrule
\multicolumn{10}{l}{\textit{CW Attack ($\ell_2$, $c=0.001$):}} \\
\hspace{1em}DiffPure-DDPM \hspace{0.2em}\cite{nie2022diffusion}   & 87.0 & 81.0 & 0.8275 & 27.6519 & 92.0 & 82.0 & 0.8365 & 26.9951 & 81.5 \\
\hspace{1em}DiffPure-DDIM \hspace{0.5em}\cite{nie2022diffusion}   & 89.0 & 75.0 & 0.8384 & 26.9642 & \underline{97.0} & 83.0 & 0.8412 & 26.9593 & 79.0 \\
\hspace{1em}DP  \hspace{4.9em}\cite{lee2023defending}   & 93.0 & 44.0 & 0.8624 & 29.5208 & 94.0 & 49.0 & 0.8744 & 29.1258 & 46.5 \\
\hspace{1em}TS  \hspace{4.55em}\cite{hwang2024temporal}   & 90.0 & 79.0 &\underline{0.9468} & 25.8049 & 91.0 & 70.0 & \underline{0.9169} & 25.5281 & 74.5 \\
\hspace{1em}FlowPure \hspace{1.15em}\cite{collaert2025flowpure}     & 93.0 & 82.0 & 0.8919 & 27.5526 & 95.0 & 79.0 & 0.8625 & 28.9614 & 80.5 \\
\hspace{1em}\textbf{FMVP$^{\mathrm{CW}}$} \hspace{2.2em}(Ours)      & 96.0 & \underline{89.0} & 0.8952 & 27.9621 & 94.0 & \underline{90.0} & 0.8697 & 29.6149 & \underline{89.5} \\
\hspace{1em}\textbf{FMVP$^{\mathrm{PGD}}$} \hspace{1.80em}(Ours)      & \underline{92.0} & 79.0 & 0.8837 & \underline{31.6549} & 92.0 & 83.0 & 0.8737 & \underline{30.8614} & 81.0 \\
\hspace{1em}\textbf{FMVP$^{\mathrm{Gaussian}}$}  \hspace{0.4em}(Ours)  & 96.0 & 83.0 & 0.9019 & 30.4552 & 94.0 & 88.0 & 0.8803 & 29.6493 & 85.5 \\

\midrule
\multicolumn{10}{l}{\textit{DiffHammer (Adaptive, $\epsilon=8/255$):}} \\
\hspace{1em}DiffPure-DDPM \hspace{0.2em}\cite{nie2022diffusion}   & 85.0 & 8.0 & 0.8519 & 28.9061 & 89.0 & 9.0 & 0.8410 & 27.9633 & 8.5 \\
\hspace{1em}DiffPure-DDIM \hspace{0.5em}\cite{nie2022diffusion}   & 87.0 & 11.0 & 0.8614 & 28.5521 & 90.0 & 13.0 & 0.8526 & 26.9667 & 12.0 \\
\hspace{1em}DP  \hspace{4.9em}\cite{lee2023defending}   & \underline{96.0} & 19.0 & 0.8692 & 28.1547 & 95.0 & 21.0 & 0.8699 & 29.4152 & 20.0 \\
\hspace{1em}TS  \hspace{4.55em}\cite{hwang2024temporal}   & 93.0 & 6.0 & \underline{0.9531} & 26.4854 & 90.0 & 5.0 & \underline{0.9452} & 25.8199 & 5.5 \\
\hspace{1em}FlowPure \hspace{1.15em}\cite{collaert2025flowpure}     & 95.0 & 12.0 & 0.8912 & 28.9106 & \underline{96.0} & 16.0 & 0.8891 & 28.0215 & 14.0 \\
\hspace{1em}\textbf{FMVP$^{\mathrm{CW}}$} \hspace{2.2em}(Ours)      & 95.0 & 16.0 & 0.8963 & \underline{28.9452} & 94.0 & 19.0 & 0.8809 & 28.9134 & 17.5 \\
\hspace{1em}\textbf{FMVP$^{\mathrm{PGD}}$} \hspace{1.80em}(Ours)      & 92.0 & \underline{20.0} & 0.8852 & 28.5159 & 93.0 & \underline{24.0} & 0.8799 & \underline{29.4937} & 22.0 \\
\hspace{1em}\textbf{FMVP$^{\mathrm{Gaussian}}$}  \hspace{0.4em}(Ours)  & 94.0 & \underline{31.0} & 0.8971 & 27.9523 & 92.0 & \underline{33.0} & 0.8762 & 28.6634 & \underline{32.0} \\
\bottomrule
\end{tabular}
\vspace{0.2cm}
\caption{Comparison of purification performance and quality under PGD, CW, and adaptive DiffHammer attacks on C3D. We report Robust  Accuracy (\textbf{Robust}, \% ($\uparrow$)), Clean Accuracy after purification (\textbf{Clean}, \% ($\uparrow$)), and video quality metrics (SSIM/PSNR) ($\uparrow$). \textbf{Avg. Robust} ($\uparrow$) denotes the average robust accuracy across both datasets. Underlined values indicate noteworthy results.}
\label{tab:main_results}
\end{table*}

\begin{figure}[t]
  \includegraphics[width=1\linewidth]{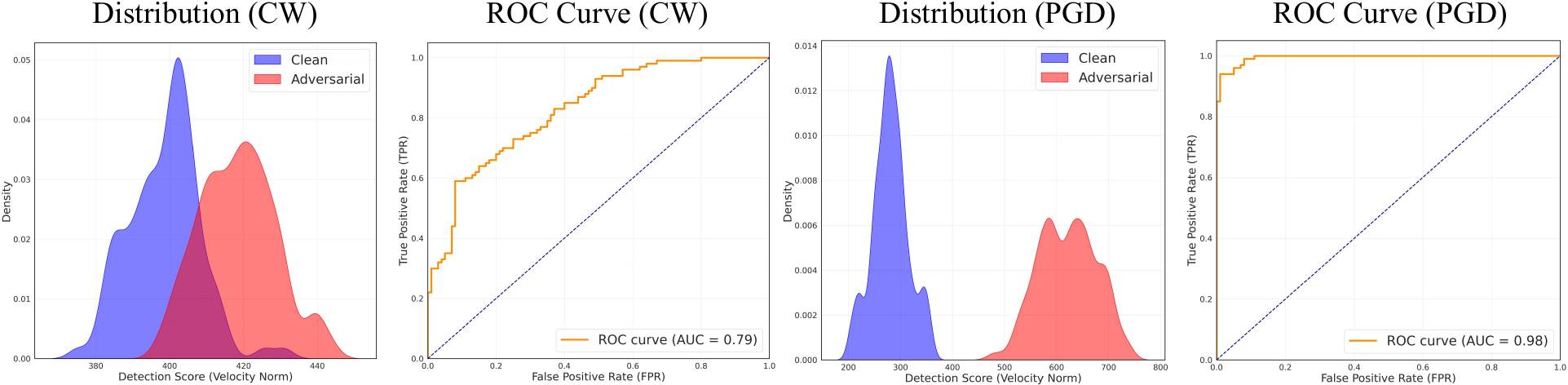}
\vspace{-3mm}
  \caption{Distribution of detection scores and ROC curves. Adversarial samples usually score higher than clean ones (horizontal axis), especially for PGD, while CW’s imperceptibility causes partial score overlap with clean samples.}
\label{figs:Detection}
\end{figure}

\section{Experiments}
\subsection{Experimental  Settings}
$\mathbf{Competitors.}$
Several state-of-the-art defense baselines are evaluated: two diffusion-based purification methods—DiffPure~\cite{nie2022diffusion} (in both DDPM and DDIM variants), Defense Patterns (DP)~\cite{lee2023defending}, Temporal Shuffling (TS)~\cite{hwang2024temporal} and FlowPure~\cite{collaert2025flowpure}. 

$\mathbf{Attack\ Methods.}$
We evaluate under three attack strategies: PGD~\cite{madry2017towards}, CW~\cite{carlini2017towards}, and the adaptive DH attack~\cite{wang2024diffhammer}, where DH is employed to probe the worst-case robustness of generative defense methods.

$\mathbf{Victim\ Models.}$
The victim models include C3D~\cite{tran2015learning}, I3D~\cite{carreira2017quo}, and R3D~\cite{tran2018closer}. C3D and I3D are used as primary targets for attacks, while R3D serves as an additional cross-model validation to assess whether the defense efficacy of FMVP depends on the specific architecture of the victim model. All models are pretrained on their respective video datasets and achieve competitive clean accuracy.

$\mathbf{Video\ Datasets.}$
Experiments are conducted on UCF-101~\cite{soomro2012ucf101} and HMDB-51~\cite{kuehne2011hmdb}, two standard benchmarks for action recognition. To train the 3D U-Net that predicts the velocity field $v_\theta(\mathbf{x}_t, t)$, we merge both datasets and randomly sample 60\% clips of the combined set as the training split. For evaluation, we randomly select 500 clean video clips from the non-overlapping held-out portion of each dataset. Videos are correctly classified by the pretrained victim models prior to attack.

$\mathbf{Evaluation\ Metrics.}$
We report two key metrics:  
(i) \textit{Robust Accuracy (Robust)} and \textit{Clean Accuracy (Clean)}, defined respectively as the proportion of adversarially misclassified samples correctly restored after purification, and the classification accuracy on clean samples after applying the defense.
(ii) \textit{Reconstruction Quality}, measured by SSIM and PSNR between the purified and original clean videos, enabling direct comparison with generation-based defenses.

\subsection{Results}

\subsubsection{$\mathbf{Performance\ against\ Standard\ Attacks}$} 
In the standard gray-box settings, our FMVP framework consistently outperforms state-of-the-art baselines in robust accuracy and holds a dominant advantage in visual fidelity. 
As shown in Table~\ref{tab:main_results}, the attack-aware variants of FMVP achieve the best robust accuracy under their corresponding attacks: FMVP$^{\mathrm{PGD}}$ against PGD (87.5\%) and FMVP$^{\mathrm{CW}}$ against CW (89.5\%). Moreover, the general FMVP$^{\mathrm{Gaussian}}$ outperforms all state-of-the-art methods, validating the synergy of our Masked Flow Matching in shattering adversarial structures and the Frequency-Gated Loss in ensuring high-fidelity semantic reconstruction, evidenced by competitive SSIM/PSNR scores. Visual results and more numerical results are provided in Section~\ref{sec:More_Results} of the Appendix.

\subsubsection{$\mathbf{Robustness\ against\ Adaptive\ Attacks}$}
Under the rigorous DiffHammer adaptive attack, which breaks most defenses via gradient approximation and Expectation Over Transformation (EOT), traditional methods like TS and DiffPure collapse to near-random performance ($<14\%$). In contrast, FMVP maintains substantial robustness, with FMVP$^{\mathrm{Gaussian}}$ achieving the best overall performance (32.0\% Avg. Robust). Crucially, the Generalist model (FMVP$^{\mathrm{Gaussian}}$) outperforms attack-specific variants, suggesting that its generalized manifold projection avoids overfitting to fixed perturbation patterns, thereby significantly complicating gradient estimation for adaptive adversaries.

\begin{figure*}[t]
\begin{center}
  \includegraphics[width=0.95\linewidth]{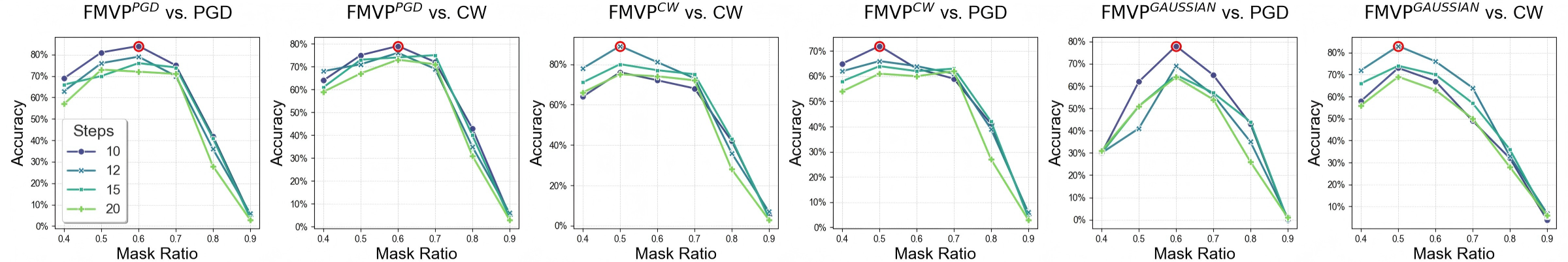}
\end{center}
\vspace{-3mm}
  \caption{Robust Accuracy vs. Masking Ratio and Euler Steps.}
\label{figs:Grid_Search}
\end{figure*}

\subsection{Ablation Study}
To validate the intrinsic contribution of each component, we conduct an ablation study on FMVP. As shown in Table~\ref{tab:ablation}, the Baseline (standard CFM with MSE) yields suboptimal robustness, suffering from ``lazy learning'' where the model fails to sufficiently dislodge inputs from the adversarial manifold. Introducing Masking significantly boosts performance by physically shattering the global coherence of adversarial patterns, forcing the model to rely on semantic inpainting. Meanwhile, the FGLoss independently improves fidelity by functioning as a soft spectral filter that suppresses high-frequency residuals, outperforming the use of LPIPS loss~\cite{zhang2018unreasonable}. Moreover, the random masking strategy offers a clear advantage in defending against adaptive attacks.

The impact of masking ratio and solver steps is illustrated in Figure~\ref{figs:Grid_Search}. The robust accuracy exhibits a consistent inverted-U trend with respect to the masking ratio, identifying an optimal range between 0.5 and 0.6 that effectively balances the destruction of adversarial patterns with the preservation of semantic content. Furthermore, the method demonstrates high inference efficiency, as fewer Euler steps (e.g., 10 and 12) consistently achieve peak performance compared to larger step counts across different settings.

\subsection{Discussion}
\begin{figure}[t]
  \includegraphics[width=0.95\linewidth]{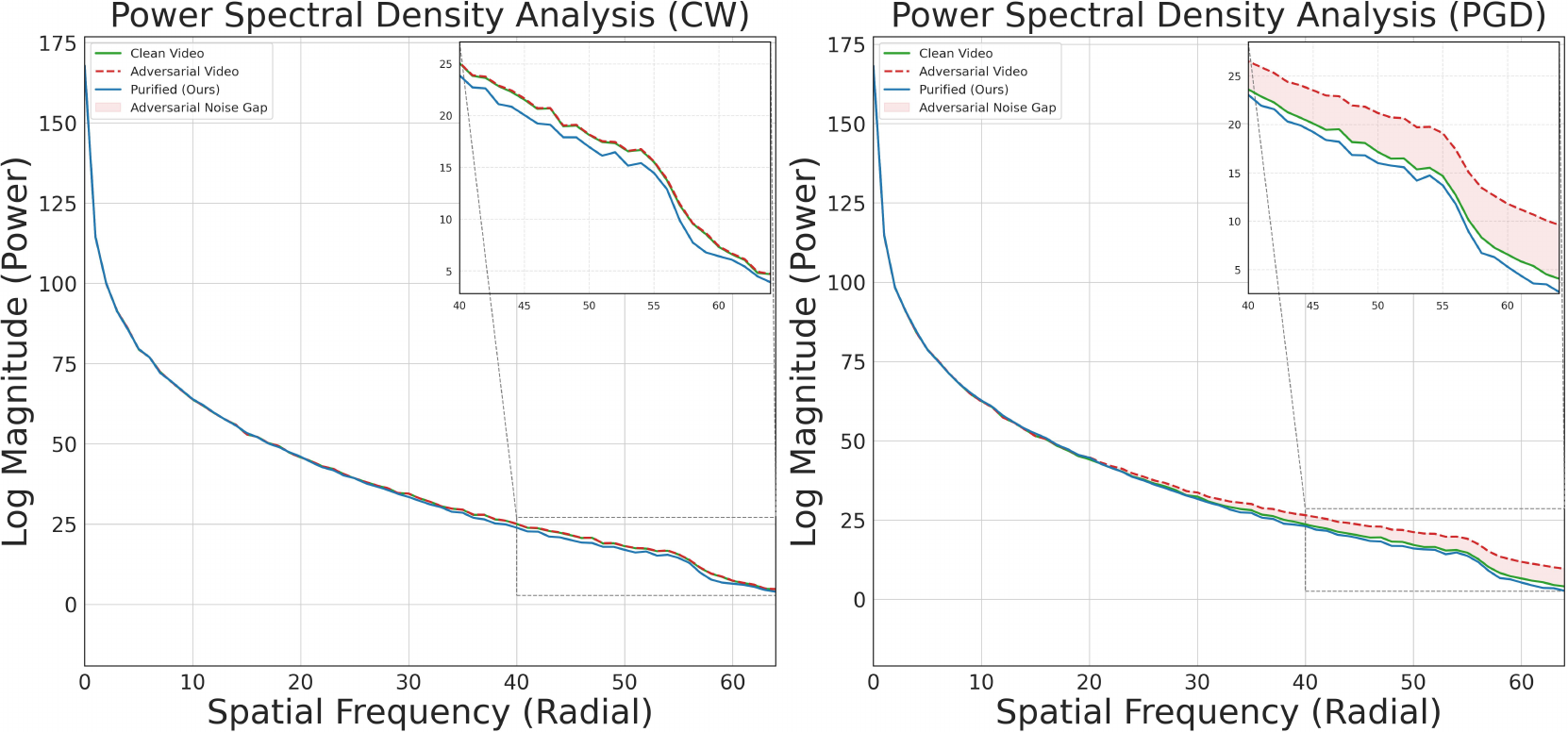}
\vspace{-3mm}
  \caption{Power Spectral Density Analysis.}
\label{figs:PSD_analysis}
\end{figure}

\subsubsection{$\mathbf{Adversarial\ Detection\ via\ Velocity\ Norms}$}
Beyond purification, FMVP naturally serves as a zero-shot adversarial detector by leveraging the kinetic properties of the learned flow. We define the detection score as the $L_2$ norm of the predicted velocity field at $t=0$, i.e., $\|\mathbf{v}_\theta(\mathbf{x}_{\text{input}}, 0)\|_2$. This metric quantifies the discrepancy between clean and adversarial inputs in terms of the $L_2$ norm of their predicted velocity fields at $t=0$.
As shown in Figure~\ref{figs:Detection}, this energy gap enables FMVP to achieve near-perfect detection against PGD attacks (AUC = 0.98). For the optimization-based CW attack, which minimizes perturbation norms to extreme levels, FMVP still retains strong discriminative power (AUC = 0.79), demonstrating the sensitivity of our frequency-gated objective to subtle off-manifold anomalies.

\subsubsection{$\mathbf{Spectral\ Analysis\ of\ Adversarial\ Purification}$}
As shown in \reffig{figs:PSD_analysis}, the Power Spectral Density (PSD) analysis reveals that PGD induces a prominent high-frequency energy surge due to explicit gradient perturbations. By leveraging the FGL Loss, FMVP effectively functions as a spectral filter to identify and suppress these anomalies. Consequently, the purified spectra (blue) closely align with the clean baselines (green) in both scenarios, validating that our method eliminates adversarial noise while preserving low-frequency semantic fidelity. The optimization-based CW attack is nearly imperceptible due to its minimal noise, but FMVP’s masking strategy effectively weakens it by disrupting the structural consistency of its perturbations.

\begin{table}[t]
\centering
\renewcommand{\arraystretch}{0.5} 
\setlength{\tabcolsep}{3pt}

\begin{tabular}{l|cc|cccc}
\toprule
\multirow{2}{*}{\textbf{Method}} & \multicolumn{2}{c|}{\textbf{Module}} & \multicolumn{4}{c}{\textbf{Acc. (\%)}} \\
\cmidrule(lr){2-3} \cmidrule(lr){4-7}
 & Mask & FGL & Clean & PGD & CW & DH\\
\midrule
Base    & $-$ & $-$ & 94.5 & 79.0 & 80.5 & 14.0 \\
+ Masking   & $\checkmark$ & $-$ & 93.0 & 84.0 & 86.0 & 24.0 \\
+ FGLoss      & $-$ & $\checkmark$ & 95.5 & 80.0 & 81.0 & 16.0 \\
+ LPIPS      & $\checkmark$ & $-$ & 94.0 & 79.0 & 78.0 & 20.0 \\
\midrule
FMVP & $\checkmark$ & $\checkmark$ & \underline{96.0} & \underline{87.5} & \underline{89.5} & \underline{32.0}\\
\bottomrule
\end{tabular}
\vspace{0.2cm}
\caption{Ablation study of the FMVP framework. Results are averaged across UCF-101 and HMDB-51.}
\label{tab:ablation}
\end{table}

\begin{table}[t]
\centering
\renewcommand{\arraystretch}{0.1}
\setlength{\tabcolsep}{10pt}
\begin{tabular}{l|c}
\toprule
\textbf{Method} & \textbf{Latency (s/video) $\downarrow$} \\
\midrule
DiffPure (DDPM)   & 35.91 \\
DiffPure (DDIM)   & 3.62 \\
\midrule
FMVP (10 steps)  & \underline{1.44} \\
FMVP (12 steps)  & 1.72 \\
FMVP (15 steps)  & 2.14 \\
FMVP (20 steps)  & 2.88 \\
\bottomrule
\end{tabular}

\vspace{0.2cm}
\caption{Inference time comparison on a single NVIDIA GeForce 4090 GPU.}
\label{tab:time_cost}
\end{table}
\subsubsection{$\mathbf{Inference\ Time\ Evaluation}$}
Table~\ref{tab:time_cost} reports the inference time of diffusion-based purification methods and FMVP with Euler solver steps of 10, 12, 15, and 20. The results show that FMVP achieves the fastest inference speed.

\section{Conclusion}
In this paper, we propose FMVP, a novel purification framework leveraging Conditional Flow Matching. By integrating a stochastic masking strategy with a Frequency-Gated Loss, FMVP effectively shatters global adversarial patterns while preserving low-frequency semantic fidelity. Extensive experiments on UCF-101 and HMDB-51 demonstrate that FMVP significantly outperforms state-of-the-art methods against both standard (PGD and CW) and strong adaptive (DH) attacks, offering superior trade-offs between robustness and efficiency. Furthermore, the intrinsic velocity properties of FMVP enable effective zero-shot adversarial detection, establishing a versatile defense solution for secure video recognition. Its high efficiency and plug-and-play nature make FMVP a practical solution for securing real-world applications, such as autonomous driving, video surveillance, and video content moderation.




\clearpage
\bibliographystyle{named}
\bibliography{ijcai26}

\begin{thebibliography}{}

\bibitem[\protect\citeauthoryear{Abdou}{2022}]{abdou2022literature}
Mohamed~A Abdou.
\newblock Literature review: Efficient deep neural networks techniques for medical image analysis.
\newblock {\em Neural Computing and Applications}, 34(8):5791--5812, 2022.

\bibitem[\protect\citeauthoryear{Athalye \bgroup \em et al.\egroup }{2018}]{athalye2018synthesizing}
Anish Athalye, Logan Engstrom, Andrew Ilyas, and Kevin Kwok.
\newblock Synthesizing robust adversarial examples.
\newblock In {\em International conference on machine learning}, pages 284--293. PMLR, 2018.

\bibitem[\protect\citeauthoryear{Carlini and Wagner}{2017}]{carlini2017towards}
Nicholas Carlini and David Wagner.
\newblock Towards evaluating the robustness of neural networks.
\newblock In {\em 2017 ieee symposium on security and privacy (sp)}, pages 39--57. Ieee, 2017.

\bibitem[\protect\citeauthoryear{Carreira and Zisserman}{2017}]{carreira2017quo}
Joao Carreira and Andrew Zisserman.
\newblock Quo vadis, action recognition? a new model and the kinetics dataset.
\newblock In {\em Proceedings of the IEEE Conference on Computer Vision and Pattern Recognition}, pages 6299--6308, 2017.

\bibitem[\protect\citeauthoryear{Chen \bgroup \em et al.\egroup }{2018}]{chen2018neural}
Ricky~TQ Chen, Yulia Rubanova, Jesse Bettencourt, and David~K Duvenaud.
\newblock Neural ordinary differential equations.
\newblock {\em Advances in neural information processing systems}, 31, 2018.

\bibitem[\protect\citeauthoryear{Collaert \bgroup \em et al.\egroup }{2025}]{collaert2025flowpure}
Elias Collaert, Abel Rodr{\'\i}guez, Sander Joos, Lieven Desmet, and Vera Rimmer.
\newblock Flowpure: Continuous normalizing flows for adversarial purification.
\newblock {\em arXiv preprint arXiv:2505.13280}, 2025.

\bibitem[\protect\citeauthoryear{Gowal \bgroup \em et al.\egroup }{2020}]{gowal2020uncovering}
Sven Gowal, Chongli Qin, Jonathan Uesato, Timothy Mann, and Pushmeet Kohli.
\newblock Uncovering the limits of adversarial training against norm-bounded adversarial examples.
\newblock {\em arXiv preprint arXiv:2010.03593}, 2020.

\bibitem[\protect\citeauthoryear{Ho \bgroup \em et al.\egroup }{2020}]{ho2020denoising}
Jonathan Ho, Ajay Jain, and Pieter Abbeel.
\newblock Denoising diffusion probabilistic models.
\newblock {\em Advances in neural information processing systems}, 33:6840--6851, 2020.

\bibitem[\protect\citeauthoryear{Hwang \bgroup \em et al.\egroup }{2024}]{hwang2024temporal}
Jaehui Hwang, Huan Zhang, Jun-Ho Choi, Cho-Jui Hsieh, and Jong-Seok Lee.
\newblock Temporal shuffling for defending deep action recognition models against adversarial attacks.
\newblock {\em Neural Networks}, 169:388--397, 2024.

\bibitem[\protect\citeauthoryear{Ji \bgroup \em et al.\egroup }{2012}]{ji20123d}
Shuiwang Ji, Wei Xu, Ming Yang, and Kai Yu.
\newblock 3d convolutional neural networks for human action recognition.
\newblock {\em IEEE Transactions on Pattern Analysis and Machine Intelligence}, 35(1):221--231, 2012.

\bibitem[\protect\citeauthoryear{Kuehne \bgroup \em et al.\egroup }{2011}]{kuehne2011hmdb}
Hildegard Kuehne, Hueihan Jhuang, Est{\'\i}baliz Garrote, Tomaso Poggio, and Thomas Serre.
\newblock Hmdb: a large video database for human motion recognition.
\newblock In {\em 2011 International conference on computer vision}, pages 2556--2563. IEEE, 2011.

\bibitem[\protect\citeauthoryear{Lee and Ro}{2023}]{lee2023defending}
Hong~Joo Lee and Yong~Man Ro.
\newblock Defending video recognition model against adversarial perturbations via defense patterns.
\newblock {\em IEEE Transactions on Dependable and Secure Computing}, 21(4):4110--4121, 2023.

\bibitem[\protect\citeauthoryear{Li \bgroup \em et al.\egroup }{2025}]{li2025towards}
Hangyu Li, Yixin Zhang, Jiangchao Yao, Nannan Wang, and Bo~Han.
\newblock Towards regularized mixture of predictions for class-imbalanced semi-supervised facial expression recognition.
\newblock In {\em Proceedings of the Thirty-Fourth International Joint Conference on Artificial Intelligence}, pages 1377--1385, 2025.

\bibitem[\protect\citeauthoryear{Lin \bgroup \em et al.\egroup }{2024}]{lin2024video}
Bin Lin, Yang Ye, Bin Zhu, Jiaxi Cui, Munan Ning, Peng Jin, and Li~Yuan.
\newblock Video-llava: Learning united visual representation by alignment before projection.
\newblock In {\em Proceedings of the 2024 conference on empirical methods in natural language processing}, pages 5971--5984, 2024.

\bibitem[\protect\citeauthoryear{Lipman \bgroup \em et al.\egroup }{2022}]{lipman2022flow}
Yaron Lipman, Ricky~TQ Chen, Heli Ben-Hamu, Maximilian Nickel, and Matt Le.
\newblock Flow matching for generative modeling.
\newblock {\em arXiv preprint arXiv:2210.02747}, 2022.

\bibitem[\protect\citeauthoryear{Liu \bgroup \em et al.\egroup }{2022}]{liu2022flow}
Xingchao Liu, Chengyue Gong, and Qiang Liu.
\newblock Flow straight and fast: Learning to generate and transfer data with rectified flow.
\newblock {\em arXiv preprint arXiv:2209.03003}, 2022.

\bibitem[\protect\citeauthoryear{Lugmayr \bgroup \em et al.\egroup }{2022}]{lugmayr2022repaint}
Andreas Lugmayr, Martin Danelljan, Andres Romero, Fisher Yu, Radu Timofte, and Luc Van~Gool.
\newblock Repaint: Inpainting using denoising diffusion probabilistic models.
\newblock In {\em Proceedings of the IEEE/CVF conference on computer vision and pattern recognition}, pages 11461--11471, 2022.

\bibitem[\protect\citeauthoryear{Madry \bgroup \em et al.\egroup }{2017}]{madry2017towards}
Aleksander Madry, Aleksandar Makelov, Ludwig Schmidt, Dimitris Tsipras, and Adrian Vladu.
\newblock Towards deep learning models resistant to adversarial attacks.
\newblock {\em arXiv preprint arXiv:1706.06083}, 2017.

\bibitem[\protect\citeauthoryear{Nie \bgroup \em et al.\egroup }{2022}]{nie2022diffusion}
Weili Nie, Brandon Guo, Yujia Huang, Chaowei Xiao, Arash Vahdat, and Anima Anandkumar.
\newblock Diffusion models for adversarial purification.
\newblock {\em arXiv preprint arXiv:2205.07460}, 2022.

\bibitem[\protect\citeauthoryear{Pouya}{2018}]{pouya2018defense}
Samangouei Pouya.
\newblock Defense-gan: Protecting classifiers against adversarial attacks using generative models.
\newblock {\em Retrieved from https://arXiv: 1805.06605}, 2018.

\bibitem[\protect\citeauthoryear{Samangouei \bgroup \em et al.\egroup }{2018}]{samangouei2018defense}
Pouya Samangouei, Maya Kabkab, and Rama Chellappa.
\newblock Defense-gan: Protecting classifiers against adversarial attacks using generative models.
\newblock {\em arXiv preprint arXiv:1805.06605}, 2018.

\bibitem[\protect\citeauthoryear{Singh \bgroup \em et al.\egroup }{2023}]{singh2023revisiting}
Naman~Deep Singh, Francesco Croce, and Matthias Hein.
\newblock Revisiting adversarial training for imagenet: Architectures, training and generalization across threat models.
\newblock {\em Advances in Neural Information Processing Systems}, 36:13931--13955, 2023.

\bibitem[\protect\citeauthoryear{Song \bgroup \em et al.\egroup }{2020}]{song2020denoising}
Jiaming Song, Chenlin Meng, and Stefano Ermon.
\newblock Denoising diffusion implicit models.
\newblock {\em arXiv preprint arXiv:2010.02502}, 2020.

\bibitem[\protect\citeauthoryear{Soomro \bgroup \em et al.\egroup }{2012}]{soomro2012ucf101}
Khurram Soomro, Amir~Roshan Zamir, and Mubarak Shah.
\newblock Ucf101: A dataset of 101 human actions classes from videos in the wild.
\newblock {\em arXiv preprint arXiv:1212.0402}, 2012.

\bibitem[\protect\citeauthoryear{Su \bgroup \em et al.\egroup }{2019}]{su2019one}
Jiawei Su, Danilo~Vasconcellos Vargas, and Kouichi Sakurai.
\newblock One pixel attack for fooling deep neural networks.
\newblock {\em IEEE Transactions on Evolutionary Computation}, 23(5):828--841, 2019.

\bibitem[\protect\citeauthoryear{Tang \bgroup \em et al.\egroup }{2025}]{tang2025video}
Yunlong Tang, Jing Bi, Siting Xu, Luchuan Song, Susan Liang, Teng Wang, Daoan Zhang, Jie An, Jingyang Lin, Rongyi Zhu, et~al.
\newblock Video understanding with large language models: A survey.
\newblock {\em IEEE Transactions on Circuits and Systems for Video Technology}, 2025.

\bibitem[\protect\citeauthoryear{Tran \bgroup \em et al.\egroup }{2015}]{tran2015learning}
Du~Tran, Lubomir Bourdev, Rob Fergus, Lorenzo Torresani, and Manohar Paluri.
\newblock Learning spatiotemporal features with 3d convolutional networks.
\newblock In {\em Proceedings of the IEEE International Conference on Computer Vision}, pages 4489--4497, 2015.

\bibitem[\protect\citeauthoryear{Tran \bgroup \em et al.\egroup }{2018}]{tran2018closer}
Du~Tran, Heng Wang, Lorenzo Torresani, Jamie Ray, Yann LeCun, and Manohar Paluri.
\newblock A closer look at spatiotemporal convolutions for action recognition.
\newblock In {\em Proceedings of the IEEE conference on Computer Vision and Pattern Recognition}, pages 6450--6459, 2018.

\bibitem[\protect\citeauthoryear{von Platen \bgroup \em et al.\egroup }{2022}]{von-platen-etal-2022-diffusers}
Patrick von Platen, Suraj Patil, Anton Lozhkov, Pedro Cuenca, Nathan Lambert, Kashif Rasul, Mishig Davaadorj, and Thomas Wolf.
\newblock Diffusers: State-of-the-art diffusion models.
\newblock \url{https://github.com/huggingface/diffusers}, 2022.

\bibitem[\protect\citeauthoryear{Wang and Deng}{2021}]{wang2021deep}
Mei Wang and Weihong Deng.
\newblock Deep face recognition: A survey.
\newblock {\em Neurocomputing}, 429:215--244, 2021.

\bibitem[\protect\citeauthoryear{Wang \bgroup \em et al.\egroup }{2023}]{wang2023better}
Zekai Wang, Tianyu Pang, Chao Du, Min Lin, Weiwei Liu, and Shuicheng Yan.
\newblock Better diffusion models further improve adversarial training.
\newblock In {\em International conference on machine learning}, pages 36246--36263. PMLR, 2023.

\bibitem[\protect\citeauthoryear{Wang \bgroup \em et al.\egroup }{2024a}]{wang2024diffhammer}
Kaibo Wang, Xiaowen Fu, Yuxuan Han, and Yang Xiang.
\newblock Diffhammer: Rethinking the robustness of diffusion-based adversarial purification.
\newblock {\em Advances in Neural Information Processing Systems}, 37:89535--89562, 2024.

\bibitem[\protect\citeauthoryear{Wang \bgroup \em et al.\egroup }{2024b}]{wang2024multimodal}
Mengmeng Wang, Jiazheng Xing, Boyuan Jiang, Jun Chen, Jianbiao Mei, Xingxing Zuo, Guang Dai, Jingdong Wang, and Yong Liu.
\newblock A multimodal, multi-task adapting framework for video action recognition.
\newblock In {\em Proceedings of the AAAI Conference on Artificial Intelligence}, volume~38, pages 5517--5525, 2024.

\bibitem[\protect\citeauthoryear{Wang \bgroup \em et al.\egroup }{2025}]{ijcai2025p60}
Chunjiang Wang, Kun Zhang, Yandong Liu, Zhiyang He, Xiaodong Tao, and S.~Kevin Zhou.
\newblock Mvp-cbm: Multi-layer visual preference-enhanced concept bottleneck model for explainable medical image classification.
\newblock In James Kwok, editor, {\em Proceedings of the Thirty-Fourth International Joint Conference on Artificial Intelligence, {IJCAI-25}}, pages 529--537. International Joint Conferences on Artificial Intelligence Organization, 8 2025.
\newblock Main Track.

\bibitem[\protect\citeauthoryear{Xu \bgroup \em et al.\egroup }{2021}]{xu2021action}
Feiyi Xu, Feng Xu, Jiucheng Xie, Chi-Man Pun, Huimin Lu, and Hao Gao.
\newblock Action recognition framework in traffic scene for autonomous driving system.
\newblock {\em IEEE Transactions on Intelligent Transportation Systems}, 23(11):22301--22311, 2021.

\bibitem[\protect\citeauthoryear{Yoon \bgroup \em et al.\egroup }{2021}]{yoon2021adversarial}
Jongmin Yoon, Sung~Ju Hwang, and Juho Lee.
\newblock Adversarial purification with score-based generative models.
\newblock In {\em International Conference on Machine Learning}, pages 12062--12072. PMLR, 2021.

\bibitem[\protect\citeauthoryear{Zhang \bgroup \em et al.\egroup }{2018}]{zhang2018unreasonable}
Richard Zhang, Phillip Isola, Alexei~A Efros, Eli Shechtman, and Oliver Wang.
\newblock The unreasonable effectiveness of deep features as a perceptual metric.
\newblock In {\em Proceedings of the IEEE conference on computer vision and pattern recognition}, pages 586--595, 2018.

\bibitem[\protect\citeauthoryear{Zhang \bgroup \em et al.\egroup }{2023}]{zhang2023video}
Hang Zhang, Xin Li, and Lidong Bing.
\newblock Video-llama: An instruction-tuned audio-visual language model for video understanding.
\newblock {\em arXiv preprint arXiv:2306.02858}, 2023.

\bibitem[\protect\citeauthoryear{Zhang \bgroup \em et al.\egroup }{2025}]{zhang2025adversarial}
Chiyu Zhang, Lu~Zhou, Xiaogang Xu, Jiafei Wu, and Zhe Liu.
\newblock Adversarial attacks of vision tasks in the past 10 years: A survey.
\newblock {\em ACM Computing Surveys}, 58(2):1--42, 2025.

\bibitem[\protect\citeauthoryear{Zhe \bgroup \em et al.\egroup }{2025}]{ijcai2025p276}
Ting Zhe, Mengya Han, Xiaoshuai Hao, Yong Luo, Zheng He, Xiantao Cai, and Jing Zhang.
\newblock Open-vocabulary fine-grained hand action detection.
\newblock In James Kwok, editor, {\em Proceedings of the Thirty-Fourth International Joint Conference on Artificial Intelligence, {IJCAI-25}}, pages 2476--2484. International Joint Conferences on Artificial Intelligence Organization, 8 2025.
\newblock Main Track.

\end{thebibliography}

\clearpage
\appendix

\section{The Algorithm of FMVP}
Alg.~\ref{alg:fmvp} presents the overall pipeline of FMVP during both training and inference. 

\begin{algorithm}[t]
\caption{FMVP: Training and Inference}
\label{alg:fmvp}
\begin{algorithmic}[1]
\INPUT Clean batch $\mathbf{x}^{\mathrm{clean}}$, Adv batch $\mathbf{x}^{\mathrm{adv}}$, Flow network $v_\theta$, Masking ratio range $[\rho_{min}, \rho_{max}]$, MSE loss weight $\lambda_{CFM}$, Frequency-Gated loss weight $\lambda_{FGL}$, Inference steps $N$.

\vspace{0.1cm}
\STATE \textbf{Stage 1: Training}
\STATE Sample $t \sim \mathcal{U}(0, 1)$, $\boldsymbol{\epsilon} \sim \mathcal{N}(\mathbf{0}, \mathbf{I})$, and $\rho \sim \mathcal{U}(\rho_{min}, \rho_{max})$.
\STATE Construct mask $\mathbf{m} \sim \text{Bernoulli}(1-\rho)$ and source $\mathbf{x}_0 \leftarrow \mathbf{m} \odot \mathbf{x}^{\mathrm{adv}} + (1 - \mathbf{m}) \odot \boldsymbol{\epsilon}$.
\STATE Set target $\mathbf{x}_1 \leftarrow \mathbf{x}^{\mathrm{clean}}$, interpolated state $\mathbf{x}_t \leftarrow (1 - t)\mathbf{x}_0 + t\mathbf{x}_1$, and target velocity $\mathbf{u}_t \leftarrow \mathbf{x}_1 - \mathbf{x}_0$.
\STATE Compute CFM loss $\mathcal{L}_{CFM} \leftarrow \|v_\theta(\mathbf{x}_t, t) - \mathbf{u}_t\|_2^2$.
\STATE Compute FG loss $\mathcal{L}_{FGL} \leftarrow \|\mathbf{W} \odot (|\text{FFT}(v_\theta(\mathbf{x}_t, t))| - |\text{FFT}(\mathbf{u}_t)|)\|_2^2$.
\STATE Update $\theta$ by minimizing $\mathcal{L}_{total} \leftarrow \lambda_{CFM}\mathcal{L}_{CFM} + \lambda_{FGL} \mathcal{L}_{FGL}$.

\vspace{0.1cm}
\STATE \textbf{Stage 2: Inference}
\STATE Sample $\mathbf{m}$ with ratio $\rho$ and $\boldsymbol{\epsilon} \sim \mathcal{N}(\mathbf{0}, \mathbf{I})$.
\STATE Initialize state $\mathbf{x}_0 = \mathbf{m} \odot (\mathbf{x}_{adv} + \xi \boldsymbol{\epsilon}) + (1 - \mathbf{m}) \odot \boldsymbol{\epsilon}$ and step size $\Delta t \leftarrow 1/N$.
\FOR{$k = 0$ \TO $N-1$}
    \STATE $\mathbf{x}_{{k+1}} = \mathbf{x}_{{k}} + v_\theta(\mathbf{x}_k, t_k) \cdot \Delta t$.
\ENDFOR
\STATE \textbf{return} Purified video ${\mathbf{x^{purified}}} \leftarrow \text{Clamp}({\mathbf{x}}_N, 0, 1)$.
\end{algorithmic}
\end{algorithm}

\section{More Experimental Results}
\label{sec:More_Results}

\subsection{Defense Performance on I3D}
\reftable{tab:main_results_i3d} presents the main experiments on I3D corresponding to those in the main paper, where FMVP still achieves the best performance, maintaining a consistent level of defense as observed in the extensive experiments conducted on C3D, and retains high video quality.

\begin{table*}[t]
\centering

\renewcommand{\arraystretch}{0.5}

\setlength{\tabcolsep}{3pt}

\begin{tabular}{l|cccc|cccc|c}
\toprule
\multirow{2}{*}{\textbf{Method}} 
  & \multicolumn{4}{c|}{\textbf{UCF-101}} 
  & \multicolumn{4}{c|}{\textbf{HMDB-51}} 
  & \multirow{2}{*}{\textbf{Avg. Robust}} \\
\cmidrule(lr){2-5} \cmidrule(lr){6-9}
  & \textbf{Clean} & \textbf{Robust} & \textbf{SSIM} & \textbf{PSNR}
  & \textbf{Clean} & \textbf{Robust} & \textbf{SSIM} & \textbf{PSNR} & \\
\midrule
\multicolumn{10}{l}{\textit{PGD Attack ($\ell_\infty$, $\epsilon=8/255$):}} \\
\hspace{1em}DiffPure-DDPM \hspace{0.2em}\cite{nie2022diffusion}   & 89.0 & 72.0 & 0.8372 & 28.6402 & 88.0 & 77.0 & 0.8409 & 29.1121 & 74.5 \\
\hspace{1em}DiffPure-DDIM \hspace{0.5em}\cite{nie2022diffusion}   & 93.0 & 74.0 & 0.8226 & 28.6145 & 92.0 & 81.0 & 0.8582 & 30.3015 & 77.5 \\
\hspace{1em}DP  \hspace{4.9em}\cite{lee2023defending}   & 94.0 & 58.0 & 0.8803 & \underline{29.9208} & 95.0 & 49.0 & 0.8562 & 29.7011 & 53.5 \\
\hspace{1em}TS  \hspace{4.55em}\cite{hwang2024temporal}   & \underline{96.0} & 76.0 & \underline{0.9302} & 25.0213 & 93.0 & 74.0 & \underline{0.9318} & 24.4255 & 75.0 \\
\hspace{1em}FlowPure \hspace{1.15em}\cite{collaert2025flowpure}     & 92.0 & 75.0 & 0.8860 & 29.1302 & 91.0 & 83.0 & 0.8755 & 29.7154 & 80.5 \\
\hspace{1em}\textbf{FMVP$^{\mathrm{CW}}$} \hspace{2.2em}(Ours)      & 94.0 & 86.0 & 0.8762 & 29.6543 & \underline{96.0} & 86.0 & 0.8824 & 30.2109 & 86.0 \\
\hspace{1em}\textbf{FMVP$^{\mathrm{PGD}}$} \hspace{1.80em}(Ours)      & 91.0 & \underline{89.0} & 0.8941 & 29.7286 & 95.0 & \underline{88.0} & 0.8857 & 29.6301 & \underline{88.5} \\
\hspace{1em}\textbf{FMVP$^{\mathrm{Gaussian}}$}  \hspace{0.4em}(Ours)  & 92.0 & 84.0 & 0.8987 & 28.2514 & 91.0 & 87.0 & 0.8926 & \underline{30.3112} & 85.5 \\

\midrule
\multicolumn{10}{l}{\textit{CW Attack ($\ell_2$, $c=0.001$):}} \\
\hspace{1em}DiffPure-DDPM \hspace{0.2em}\cite{nie2022diffusion}   & 90.0 & 83.0 & 0.8399 & 27.7654 & 94.0 & 81.0 & 0.8560 & 27.0912 & 82.0 \\
\hspace{1em}DiffPure-DDIM \hspace{0.5em}\cite{nie2022diffusion}   & 88.0 & 77.0 & 0.8462 & 27.0506 & \underline{96.0} & 83.0 & 0.8516 & 27.0425 & 80.0 \\
\hspace{1em}DP  \hspace{4.9em}\cite{lee2023defending}   & 93.0 & 45.0 & 0.8568 & 28.7153 & 93.0 & 58.0 & 0.8889 & 29.2104 & 51.5 \\
\hspace{1em}TS  \hspace{4.55em}\cite{hwang2024temporal}   & \underline{95.0} & 81.0 & \underline{0.9307} & 25.2002 & 94.0 & 73.0 & \underline{0.9266} & 25.6002 & 77.0 \\
\hspace{1em}FlowPure \hspace{1.15em}\cite{collaert2025flowpure}     & 91.0 & 82.0 & 0.8897 & 27.6061 & 95.0 & 84.0 & 0.8589 & 29.0427 & 83.0 \\
\hspace{1em}\textbf{FMVP$^{\mathrm{CW}}$} \hspace{2.2em}(Ours)      & 94.0 & \underline{91.0} & 0.8917 & 28.9285 & \underline{96.0} & \underline{89.0} & 0.8741 & 29.7103 & \underline{90.0} \\
\hspace{1em}\textbf{FMVP$^{\mathrm{PGD}}$} \hspace{1.80em}(Ours)      & 92.0 & 81.0 & 0.8792 & \underline{31.2930} & 95.0 & 85.0 & 0.8891 & \underline{30.9322} & 83.0 \\
\hspace{1em}\textbf{FMVP$^{\mathrm{Gaussian}}$}  \hspace{0.4em}(Ours)  & 93.0 & 85.0 & 0.9024 & 30.5001 & 95.0 & 87.0 & 0.8898 & 29.7051 & 86.0 \\

\midrule
\multicolumn{10}{l}{\textit{DiffHammer (Adaptive, $\epsilon=8/255$):}} \\
\hspace{1em}DiffPure-DDPM \hspace{0.2em}\cite{nie2022diffusion}   & 92.0 & 7.0 & 0.8862 & 28.9012 & 95.0 & 8.0 & 0.8450 & 28.0441 & 8.5 \\
\hspace{1em}DiffPure-DDIM \hspace{0.5em}\cite{nie2022diffusion}   & 91.0 & 5.0 & 0.8608 & 28.6374 & 94.0 & 11.0 & 0.8669 & 27.1501 & 8.0 \\
\hspace{1em}DP  \hspace{4.9em}\cite{lee2023defending}   & \underline{96.0} & 18.0 & 0.8736 & 27.2005 & 91.0 & 24.0 & 0.8643 & 28.5996 & 23.0 \\
\hspace{1em}TS  \hspace{4.55em}\cite{hwang2024temporal}   & 93.0 & 3.0 & \underline{0.9408} & 26.0019 & 93.0 & 7.0 & \underline{0.9268} & 26.0032 & 5.0 \\
\hspace{1em}FlowPure \hspace{1.15em}\cite{collaert2025flowpure}     & 94.0 & 11.0 & 0.8857 & 28.9901 & \underline{97.0} & 14.0 & 0.8936 & 29.1165 & 12.5 \\
\hspace{1em}\textbf{FMVP$^{\mathrm{CW}}$} \hspace{2.2em}(Ours)      & 93.0 & \underline{19.0} & 0.9108 & \underline{29.7784} & 95.0 & \underline{21.0} & 0.8863 & 28.9076 & 20.0 \\
\hspace{1em}\textbf{FMVP$^{\mathrm{PGD}}$} \hspace{1.80em}(Ours)      & \underline{96.0} & 21.0 & 0.8807 & 28.5962 & 93.0 & 20.0 & 0.8873 & \underline{29.5071} & 20.5 \\
\hspace{1em}\textbf{FMVP$^{\mathrm{Gaussian}}$}  \hspace{0.4em}(Ours)  & 94.0 & \underline{28.0} & 0.8916 & 29.0317 & 94.0 & \underline{31.0} & 0.8906 & 28.9491 & \underline{29.5} \\
\bottomrule
\end{tabular}
\vspace{0.2cm}
\caption{Comparison of purification performance and quality under PGD, CW, and adaptive DiffHammer attacks on I3D. We report Robust  Accuracy (\textbf{Robust}, \% ($\uparrow$)), Clean Accuracy after purification (\textbf{Clean}, \% ($\uparrow$)), and video quality metrics (SSIM/PSNR) ($\uparrow$). \textbf{Avg. Robust} ($\uparrow$) denotes the average robust accuracy across both datasets. Underlined values indicate noteworthy results.}
\label{tab:main_results_i3d}
\end{table*}

\subsection{Visual Results}
\begin{figure}[t]
\begin{center}
  \includegraphics[width=1\linewidth]{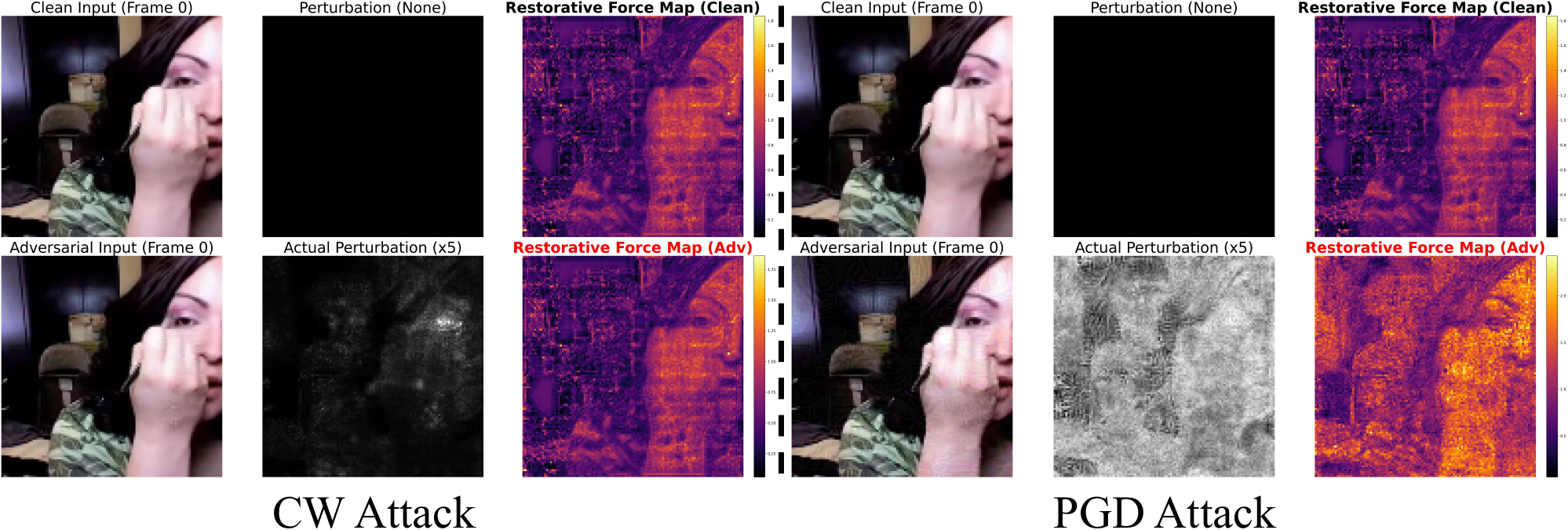}
\end{center}
\vspace{-3mm}
  \caption{Visualization of Restorative Velocity Map.}
\label{figs:visual_velocity}
\end{figure}
\subsubsection{Visual Results of Velocity Fields.}
Figure~\ref{figs:visual_velocity} shows that FMVP accurately localizes adversarial perturbations, producing strong velocity fields only where restoration is needed and preserving clean-region semantics.

\subsubsection{Comparision Visual Results with Competitors.}
\begin{figure*}[h]
\begin{center}
  \includegraphics[width=0.95\linewidth]{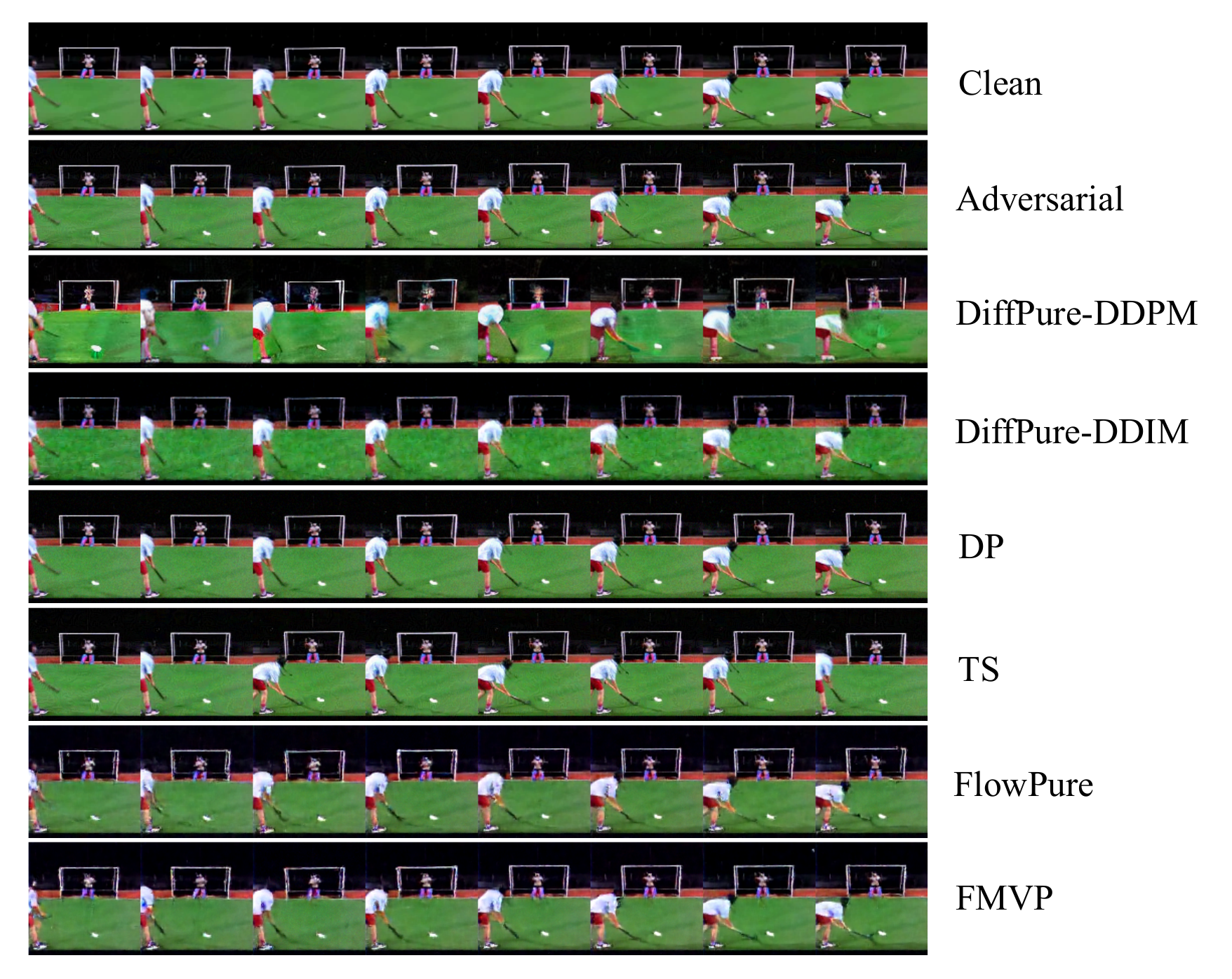}
\end{center}
\vspace{-3mm}
  \caption{Comparison of purification results across different methods: clean represents the original video, and adversarial represents the video after adversarial attack.}
\label{figs:Comparison}
\end{figure*}

Figure~\ref{figs:Comparison} compares the purification results of FMVP against other methods. Visually, FMVP outperforms both DDPM and DDIM, and achieves visual quality comparable to the generative method FlowPure. However, as shown in Table~\ref{tab:main_results}, FMVP consistently surpasses FlowPure in robust accuracy, and notably, FlowPure exhibits significantly weaker robustness against adaptive attacks compared to FMVP. Among non-generative approaches, DP introduces dense noise artifacts, while TS suffers from temporal inconsistencies due to frame-swapping, leading to logical errors in video dynamics.

\subsubsection{Comparision Visual Results of FMVP.}
\begin{figure*}[t]
\begin{center}
  \includegraphics[width=0.85\linewidth]{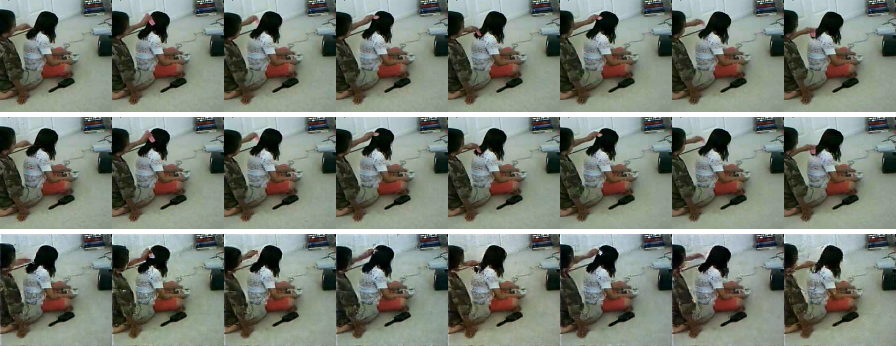}
\end{center}
\vspace{-3mm}
  \caption{HMDB-51 under CW attack: clean video, adversarial video, and purified video (from top to bottom).}
\label{figs:HMDB51_CW}
\end{figure*}

\begin{figure*}[t]
\begin{center}
  \includegraphics[width=0.85\linewidth]{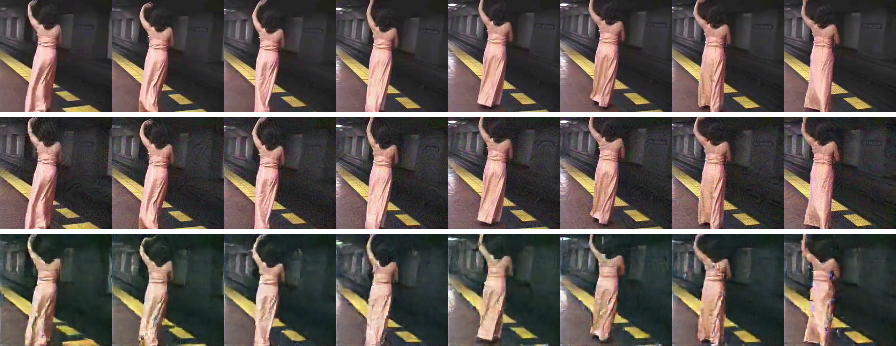}
\end{center}
\vspace{-3mm}
  \caption{HMDB-51 under PGD attack: clean video, adversarial video, and purified video (from top to bottom).}
\label{figs:HMDB51_PGD}
\end{figure*}

\begin{figure*}[t]
\begin{center}
  \includegraphics[width=0.85\linewidth]{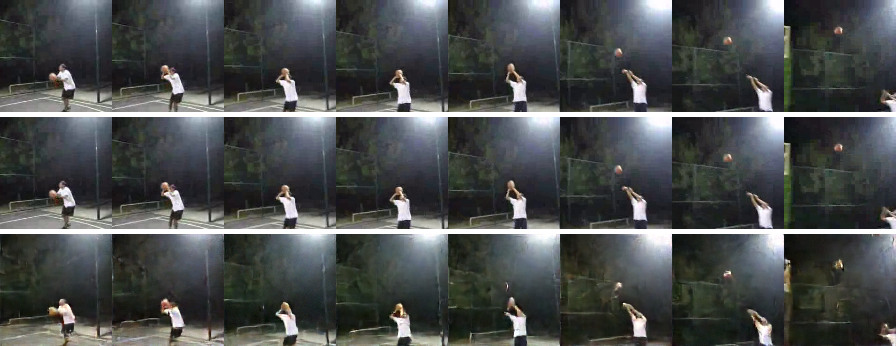}
\end{center}
\vspace{-3mm}
  \caption{UCF-101 under CW attack: clean video, adversarial video, and purified video (from top to bottom).}
\label{figs:UCF101_CW}
\end{figure*}

\begin{figure*}[t]
\begin{center}
  \includegraphics[width=0.85\linewidth]{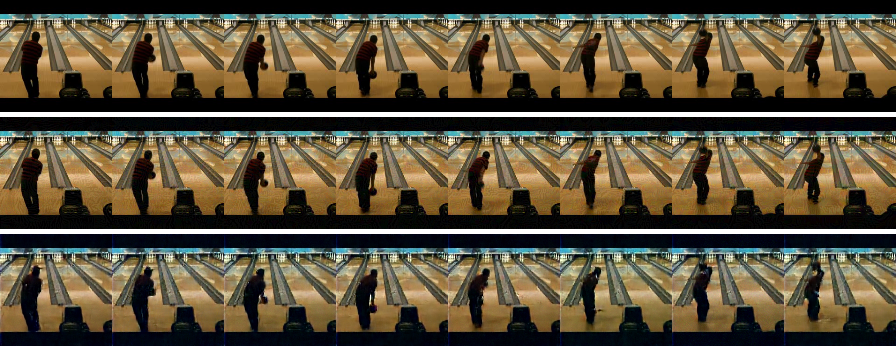}
\end{center}
\vspace{-3mm}
  \caption{UCF-101 under PGD attack: clean video, adversarial video, and purified video (from top to bottom).}
\label{figs:UCF101_PGD}
\end{figure*}

\reffig{figs:HMDB51_CW}, \reffig{figs:HMDB51_PGD}, \reffig{figs:UCF101_CW} and \reffig{figs:UCF101_PGD} show additional visual results of FMVP\(^{\text{Gaussian}}\) on HMDB-51 and UCF-101 under CW and PGD attacks. Each figure displays (top to bottom): clean, adversarial, and purified videos. The purified outputs appear highly natural, demonstrating effective disruption of adversarial patterns with faithful semantic reconstruction.

\subsection{Visualizing the Purification Trajectory}
\begin{figure*}[t]
\begin{center}
  \includegraphics[width=0.85\linewidth]{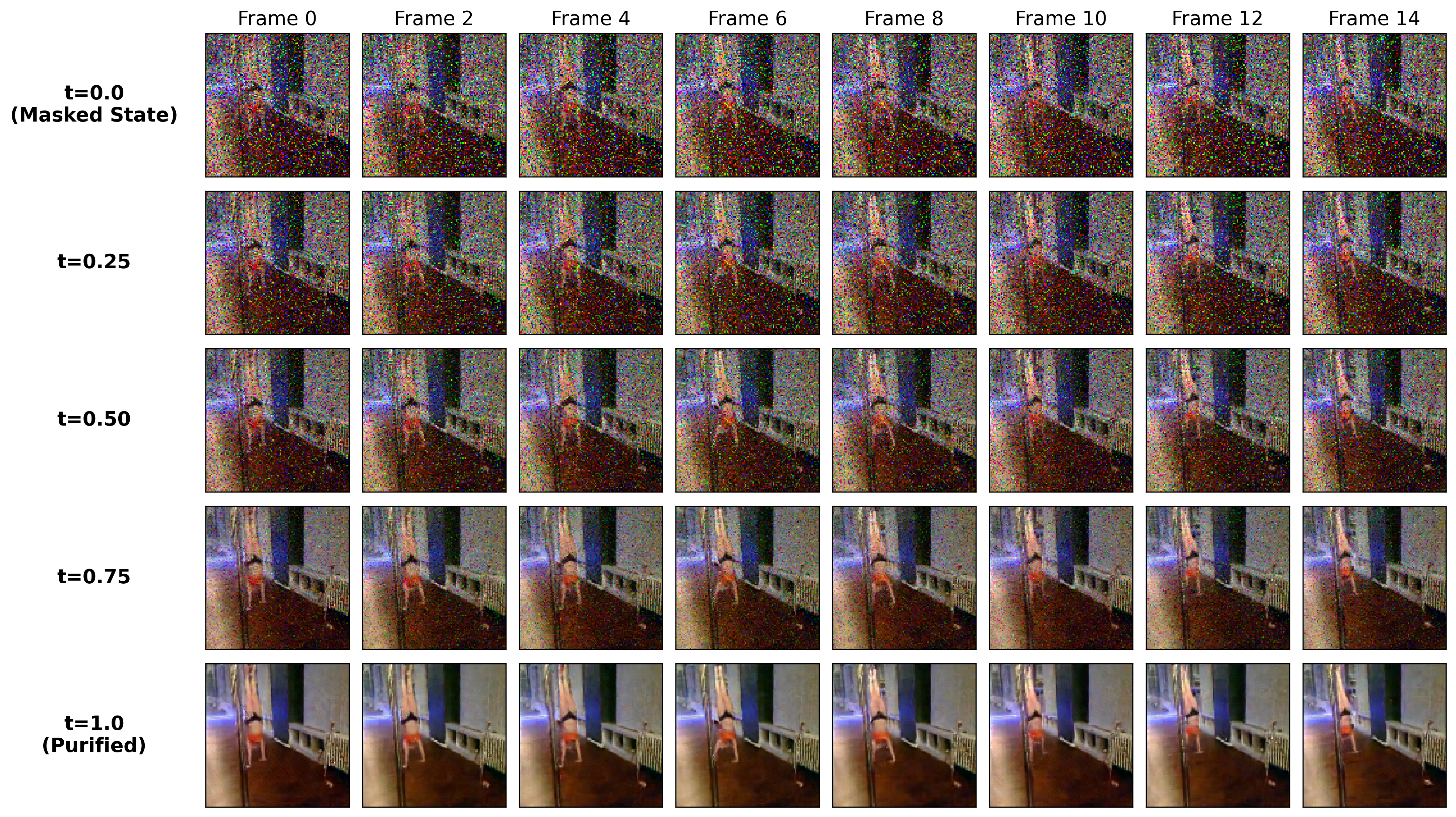}
\end{center}
\vspace{-3mm}
  \caption{Reconstruction trajectory of FMVP on HMDB-51: Gaussian noise is filled into mask regions that disrupt adversarial patterns, and the purification process is visualized over 10 Euler steps from \( t=0 \) to \( t=1 \).}
\label{figs:trajectory_vis_hmdb}
\end{figure*}

\begin{figure*}[t]
\begin{center}
  \includegraphics[width=0.85\linewidth]{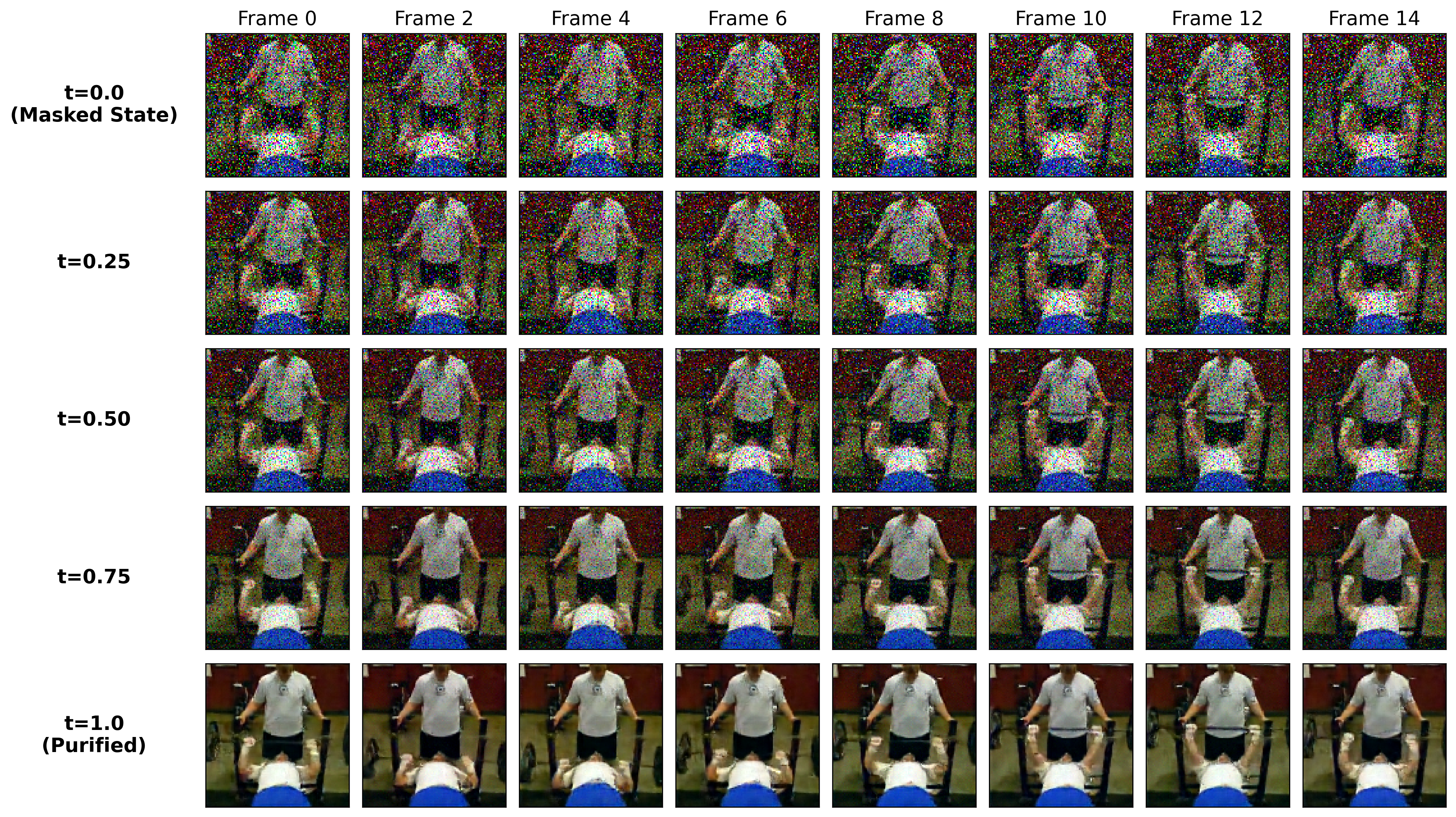}
\end{center}
\vspace{-3mm}
  \caption{Reconstruction trajectory of FMVP on UCF-101: Gaussian noise is filled into mask regions that disrupt adversarial patterns, and the purification process is visualized over 10 Euler steps from \( t=0 \) to \( t=1 \).}
\label{figs:trajectory_vis_ucf}
\end{figure*}

We also visualize the reconstruction process of FMVP. \reffig{figs:trajectory_vis_hmdb} and \reffig{figs:trajectory_vis_ucf} show sample videos from UCF-101 and HMDB-51, respectively, where Gaussian noise is filled into the mask regions that disrupt the adversarial pattern, and the purification results across 10 Euler steps from \( t=0 \) to \( t=1 \) are displayed.

\subsection{Cross-model transferability of defense}
\reftable{tab:cross_model_PGD} and \reftable{tab:cross_model_CW} report the performance of defense transferability, primarily to verify whether the velocity field prediction trained under a specific attack version of FMVP relies on adversarial samples generated by that same attack. Specifically, we generate adversarial examples and train the purifier on the \textbf{Source Model}, then evaluate the robust accuracy (\%) on different \textbf{Target Models}. ``Source = Target'' indicates the standard white-box defense setting, while ``Source $\neq$ Target'' indicates the black-box transfer defense setting. FMVP demonstrates strong generalization across different video backbone architectures. FMVP’s defense does not rely on adversarial examples generated from the same model architecture to achieve strong performance. Its masking mechanism effectively disrupts adversarial patterns originating from any victim model, and FGL is capable of capturing and suppressing the underlying structure of adversarial perturbations, thereby consistently maintaining comparable robustness across cross-model settings.

\begin{table}[t]
\centering
\renewcommand{\arraystretch}{1.2} 
\setlength{\tabcolsep}{4pt}       

\begin{tabular}{c|l|c|c|c}
\toprule
\multirow{2}{*}{\textbf{Source Model}} & \multirow{2}{*}{\textbf{Method}} & \multicolumn{3}{c}{\textbf{Transfer Model}} \\
\cmidrule(lr){3-5}
 & & \textbf{C3D} & \textbf{I3D} & \textbf{R3D} \\
\midrule

\multirow{2}{*}{\textbf{C3D}} 
 & FMVP$^{\mathrm{PGD}}$   & 87.5 & 86.0 & 89.0 \\ 
 & FMVP$^{\mathrm{CW}}$    & 78.5 & 80.0 & 77.5 \\
\midrule

\multirow{2}{*}{\textbf{I3D}} 
 & FMVP$^{\mathrm{PGD}}$   & 89.0 & 88.5 & 85.0 \\
 & FMVP$^{\mathrm{CW}}$    & 82.0 & 86.0 & 89.0 \\
\bottomrule
\end{tabular}
\vspace{0.2cm}
\caption{Robust Accuracy (Robust) of Cross-Model Defense Transferability Against PGD Attack}
\label{tab:cross_model_PGD}
\end{table}

\begin{table}[t]
\centering
\renewcommand{\arraystretch}{1.2}
\setlength{\tabcolsep}{4pt}       

\begin{tabular}{c|l|c|c|c}
\toprule
\multirow{2}{*}{\textbf{Source Model}} & \multirow{2}{*}{\textbf{Method}} & \multicolumn{3}{c}{\textbf{Transfer Model}} \\
\cmidrule(lr){3-5}
 & & \textbf{C3D} & \textbf{I3D} & \textbf{R3D} \\
\midrule
\multirow{2}{*}{\textbf{C3D}} 
 & FMVP$^{\mathrm{PGD}}$   & 81.0 & 83.5 & 82.0 \\ 
 & FMVP$^{\mathrm{CW}}$    & 89.5 & 90.0 & 87.5 \\
\midrule
\multirow{2}{*}{\textbf{I3D}} 
 & FMVP$^{\mathrm{PGD}}$   & 85.0 & 83.0 & 85.5 \\
 & FMVP$^{\mathrm{CW}}$    & 88.0 & 90.0 & 87.5 \\
\bottomrule
\end{tabular}
\vspace{0.2cm}
\caption{Robust Accuracy (Robust) of Cross-Model Defense Transferability Against CW Attack}
\label{tab:cross_model_CW}
\end{table}

\section{Implementation Details}

\subsection{Attacks Implementation}
We evaluate the robustness of FMVP using three distinct attack protocols with the following specific settings:
\begin{itemize}
    \item \textbf{PGD:} We employ the standard $L_\infty$ Projected Gradient Descent attack with a perturbation budget $\epsilon = 8/255$, step size $\eta = 2/255$, and number of iterations $N=10$.
    \item \textbf{CW:} For the optimization-based Carlini \& Wagner ($L_2$) attack, we perform 9 binary search steps for the constant $c$ (initialized at $10^{-3}$), with a learning rate of $0.01$, confidence $\kappa=0$, and $50$ optimization iterations per search step to align with the settings of FlowPure ~\cite{collaert2025flowpure}.
    \item \textbf{DiffHammer (DH):} For this strong adaptive white-box attack, we set the $L_\infty$ budget $\epsilon = 8/255$, step size $\alpha = 0.007$, and iterations $N=50$. To effectively estimate gradients through the stochastic masking process, we utilize Expectation Over Transformation (EOT) \cite{athalye2018synthesizing} with 5 samples per step and perform up to 3 random restarts. To manage GPU memory during backpropagation, we use reduced purification steps ($T_{grad}=4$) for gradient calculation, while the final evaluation uses the standard inference setting ($T_{eval}=10$).
\end{itemize}

\subsection{Training Settings of FMVP}
We implement FMVP using PyTorch. The velocity field estimator $v_\theta$ is instantiated as a 3D U-Net~\cite{von-platen-etal-2022-diffusers} derived from the Diffusers library, modified to accept video tensors of shape $16 \times 112 \times 112$ (Frames $\times$ Height $\times$ Width). The model is optimized using the AdamW optimizer with a learning rate of $1 \times 10^{-4}$ and a batch size of 1. Train each variant for three epochs.

Regarding the loss hyperparameters, we set the weight for the Frequency-Gated Loss as $\lambda_{FGL} = 0.2$, balancing spatial reconstruction and spectral consistency. The masking ratio $\gamma$ is dynamically sampled from a uniform distribution $\mathcal{U}(0.2, 0.6)$ during training to enforce robustness against varying corruption levels. All experiments are conducted on a single NVIDIA RTX 4090 GPU.

\end{document}